\documentclass[lettersize,journal]{IEEEtran}
\usepackage{amsmath,amsfonts}
\usepackage{algorithmic}
\usepackage{algorithm}
\usepackage{array}
\usepackage[caption=false,font=normalsize,labelfont=sf,textfont=sf]{subfig}
\usepackage{textcomp}
\usepackage{stfloats}
\usepackage{url}
\usepackage{verbatim}
\usepackage{graphicx}
\usepackage{cite}
\usepackage{color}
\usepackage{multirow}
\usepackage{booktabs}
\usepackage{makecell}
\usepackage{xcolor}
\usepackage{siunitx}  
\begin{document}

\title{Satellite Streaming Video QoE Prediction: A Real-World Subjective Database and Network-Level Prediction Models}

\author{Bowen Chen, Zaixi Shang, Jae Won Chung, David Lerner, Werner Robitza, Rakesh Rao Ramachandra Rao, Alexander Raake, and Alan C. Bovik,~\IEEEmembership{Life Fellow,~IEEE,}
\thanks{This paper was produced by the IEEE Publication Technology Group. They are in Piscataway, NJ.}
\thanks{Manuscript received April 19, 2021; revised August 16, 2021.}}

\newcommand{\cWR}[1]{\textbf{\textcolor{magenta}{Comment WR: #1}}}
\newcommand{\mWR}[1]{\textbf{\textcolor{blue}{WR: #1}}}
\newcommand{\mBC}[1]{\textbf{\textcolor{red}{BC: #1}}}
\newcommand{\cBC}[1]{\textbf{\textcolor{orange}{BC: #1}}}
\newcommand{\cAR}[1]{\textbf{\textcolor{magenta}{Comment AR: #1}}}
\newcommand{\mAR}[1]{\textbf{\textcolor{blue}{AR: #1}}}
\newcommand{\cRR}[1]{\textbf{\textcolor{green}{Comment RR: #1}}}
\newcommand{\mRR}[1]{\textbf{\textcolor{yellow}{RR: #1}}}

\markboth{Journal of \LaTeX\ Class Files,~Vol.~14, No.~8, August~2021}%
{Shell \MakeLowercase{\textit{et al.}}: A Sample Article Using IEEEtran.cls for IEEE Journals}


\maketitle

\begin{abstract}

Demand for streaming services, including satellite, continues to exhibit unprecedented growth. As subscribers increasingly demand seamless streaming, Internet Service Providers (ISPs) find themselves at the crossroads of technological advancements and rising customer expectations. To stay relevant and competitive, these ISPs must ensure their networks deliver optimal video streaming quality, a key determinant of user satisfaction. Towards this end, it is important to have accurate Quality of Experience (QoE) prediction models in place. These models can be used to optimize streaming protocols and ensure uninterrupted viewing experiences. However, achieving robust performance by these models requires extensive data sets labeled by subjective opinion scores on videos impaired by diverse playback disruptions. To bridge this data gap, we introduce the LIVE-Viasat Real-World Satellite QoE Database. This database consists of 179 videos recorded from real-world streaming services affected by various authentic distortion patterns. We also conducted a comprehensive subjective study involving 54 participants, who contributed both continuous-time opinion scores and endpoint (retrospective) QoE scores. Our analysis sheds light on various determinants influencing subjective QoE, such as stall events, spatial resolutions, bitrate, and certain network parameters. We demonstrate the usefulness of this unique new resource by evaluating the efficacy of prevalent QoE-prediction models on it. We also created a new model that maps the network parameters to predicted human perception scores, which can be used by ISPs to optimize the video streaming quality of their networks. Our proposed model, which we call SatQA, is able to accurately predict QoE using only network parameters, without any access to pixel data or video-specific metadata. estimated by Spearman's Rank Order Correlation Coefficient (SROCC), Pearson Linear Correlation Coefficient (PLCC), and Root Mean Squared Error (RMSE), indicating high accuracy and reliability. The publicly available database and code for SatQA can be found at \url{https://live.ece.utexas.edu/research/Quality/index.htm}.
\end{abstract}

\begin{IEEEkeywords}
Quality of experience, subjective video quality assessment, objective QoE model
\end{IEEEkeywords}

\section{INTRODUCTION}
Video streaming has become a vital part of daily life in the connected digital world. Human viewers worldwide consume video content in ever-increasing volumes, whether it be on entertainment platforms like Netflix, Amazon Prime, or YouTube, educational platforms like Coursera, Udemy, and edX, live streaming features on social media and gaming channels like Facebook and Twitch, and video teleconferencing on platforms like Zoom and Teams. According to Cisco, 82\% of all internet traffic, including satellite traffic, was made up of video downloads and streaming in 2022\cite{barnett2018cisco}. To ensure uninterrupted transmission of content to end users, ISPs have found themselves at the forefront of technology developments driving this video streaming revolution.

The smoothness of satellite video streaming is a complex topic that depends on both internal and external elements. Although internal characteristics, such as video encoding and compression methods influence the streaming experience, networks frequently have the biggest impact. For instance, if the bandwidth at a given time is insufficient to support a video's bitrate, particularly during periods of high traffic, deteriorated video quality and/or rebuffering events may occur. Another problem is packet loss, which occurs when data packets are misplaced while being transmitted over a satellite network because of congestion, signal deterioration, or transmission errors. Such losses may lead to video alterations, artifacts, or frame drops, depending on the choice of streaming technology. The role of network latency and jitter is also crucial. High latency can cause extensive initial buffering, while jitter, or variable latency, can interfere with the regular delivery of packets and cause sporadic buffering during viewing.

While Video Quality Assessment (VQA) models focus on aspects like encoding quality or scaling, Quality of Experience (QoE) models attempt to predict the overall degree of satisfaction or annoyance\cite{itu-t} of end-users while viewing streaming videos. Generally, QoE models seek to account for such variables as video quality, rebuffering events, latency, and others. Understanding and being able to measure users' QoE is crucial for ISPs to ensure customer satisfaction and lower churn rates toward achieving competitive advantages in a crowded streaming market.

Media streaming services that stream over satellite networks like YouTube and Netflix frequently employ HTTP-based adaptive streaming (HAS) protocols such as Dynamic Adaptive Streaming over HTTP (DASH)\cite{ref1} and HTTP Live Streaming (HLS)\cite{ref2}. These protocols divide videos into fixed-duration segments, where each segment is encoded at multiple bitrates and resolutions. The client autonomously determines the optimal bitrate for the next segment based on network conditions, buffer size, and other factors, to optimize QoE for the end user. Consequently, accurately modeling and predicting subjective streaming video QoE is a critical goal for satellite streaming service providers.


There are significant challenges in finding ways to accurately assess perceptual QoE. Perhaps the biggest hurdle is that nearly all satellite-streamed video is encrypted using streaming protocols such as HTTPS. Encryption prevents ISPs from being able to inspect packet contents, thereby concealing essential QoE information. While media streaming service providers like YouTube and Netflix employ their own video quality and QoE protocols, which include data sent from the streaming clients to the server, satellite ISPs do not have access to this information, and must resort to other methods of measuring and monitoring users' experience.

Numerous video QoE models have been devised to perceptually improve video streaming. Common measurements on videos made by these models include characteristics such as stalling events, initial buffering time, adaptive bitrate changes, and spatial and temporal perceptual distortion measurements\cite{ref3, ref4, ref5, ref6}. Exemplar pixel-domain and frame-domain video QoE models include those described in\cite{ref7, ref8, ref9, ref10, ref11, ref12, ref13, ref14, ref15, ref16, ref17, ref18,ref19}. Some models instead analyze network-level measurements to estimate pixel/frame-wise QoE model responses or user behavior using statistical tests or machine learning\cite{ref20,ref21,ref22,ref23,ref24,ref25,ref26}.



However, these existing algorithms exhibit several drawbacks. While most were trained and tested on synthetic datasets with simulated defects like stalling events or bitrate changes, real-world scenarios, with ever-changing network conditions, devices, and user expectations, present complexities that simulations might not fully capture. To enhance the design of perceptual QoE predictors, both accurate simulations and access to real-world datasets are essential. Such datasets, capturing a variety of scenarios faced by users, are invaluable for validating and refining simulated models.

To be able to design more accurate and reliable perceptual QoE predictors, databases representing real-world distortions are crucial. Toward making progress on filling some of these gaps, we introduce the LIVE-Viasat Real-World Satellite QoE Database. This new resource was designed to closely emulate real-world satellite streaming scenarios. It captures different types of network-induced distortions, including actual stalls, bitrate changes, and resolution changes, on a wide variety of video contents, capturing scenarios that users typically experience.

LIVE-Viasat Real-World Satellite QoE Database is the first subjective video database we are aware of that was constructed using network data that was drawn from a real-world streaming service. It models stalling patterns that vary by relative times and frequencies, as well as resolution changes, bitrate changes, and network parameters. On videos naturally impaired by these conditions, we conducted a human study whereby we gathered both continuous-time and endpoint (retrospective) QoE scores. It is our hope that introducing this highly realistic subjective dataset will help pave the way to the design of more accurate and user-centric satellite video QoE prediction models.

In our increasingly digital, interconnected world, QoE is an essential determinant of the success of video streaming services. Yet, there is still an unmet need for systematic understanding, modeling, and prediction of video QoE. Toward this goal, we make the following contributions:
\begin{enumerate}
     \item \textbf{LIVE-Viasat Real-world Satellite QoE Database:} We introduce the LIVE-Viasat Real-World Satellite QoE Database, which is based on a collection of 179 videos that were captured over a real satellite network. The overall dataset consists of the video recordings, the network packets that delivered the videos, QoE measurements and perception scores recorded by human subjects on all the streamed videos. We document the process of curating this set in Section III.

     \item \textbf{Subjective Study Design and Execution:} Understanding user behavior requires user feedback collection. To this end, we carried out an extensive subjective study, whereby we captured real-time subjective opinions of video QoE as well as endpoint (retrospective) subjective opinions of overall video QoE on all the collected contents. The details of this study, including the satellite set-up, the experimental protocol, and the subject demographics, are explained in Section IV. Post-processing and analysis of the subjective data are detailed in Section V.

     \item \textbf{Analysis of Perceived QoE:} Perceived QoE is not static—it dynamically evolves as users experience the videos as they play over time\cite{weiss2014temporal}. Thus we capture and analyze temporal variations of reported QoE, as well as overall QoE, identifying how factors like stalls, resolution changes, bitrate changes, video content, and network parameters influence QoE perception. Our findings are systematically presented in Section VI.

     \item \textbf{Testing Existing QoE Models:} To ascertain the robustness and utility of our database, we tested several established overall QoE prediction models. We also create a new QoE regression model that predicts overall QoE using only readily available network parameters but without access to pixel/frame data or even compressed format video data. In this way, we are able to demonstrate the usefulness and versatility of the new LIVE-Viasat Real-World Satellite QoE database, while enabling the development of advanced QoE prediction models, as discussed in Section VII.
 \end{enumerate}

This paper is developed from a conference paper\cite{icip}. This paper includes additional details on the subjective study, analysis of subjective human perception scores, the evaluation of the existing QoE models, and the newly proposed network-level QoE model.
\section{RELATED WORK}

The transition from traditional broadcast television to online video streaming has engendered an array of challenges for content creators and distributors. The once paramount focus on traditional video quality – emphasizing attributes like resolution, frame rate, and compression – now includes the desire for more encompassing measurements of users' overall Quality of Experience (QoE). While the quality of a video was once defined by its resolution and clarity in the days of broadcast TV and DVDs, with the advent of online all-digital streaming platforms like YouTube, Netflix, and Hulu, user expectations evolved along with changing bandwidth conditions and limitations, and especially those of satellite and wireless Internet.


Subjective QoE databases are the basic tools of QoE modeling, providing ground truth data upon which objective QoE prediction models can be built and validated. The first generation of QoE databases was primarily focused on generating synthetically distorted videos. These distortions were designed to emulate common streaming problems and distortions arising from temporal network fluctuations towards understanding their impact on user satisfaction\cite{ref7,ref8,ref9,ref10,ref11,ref12,ref13}. Further evolutions sought to more closely mimic real-world streaming conditions. For example, WaterlooSQoE-III\cite{ref15} and WaterlooSQoE-IV\cite{ref16} explored adaptive bitrate steaming, while LIVE-NFLX-II\cite{ref17} focused on the challenges of low bandwidth streaming. These kinds of databases make it possible to compare existing QoE models and to create QoE algorithms.

In today's content delivery landscape, where 80\% of the bits are of pictures and videos transmitted over increasingly crowded networks, ensuring optimal QoE in video streaming is becoming increasingly essential. SSIM\cite{ref3}, MS-SSIM\cite{ref4}, ST-RRED\cite{ref5}, and VMAF\cite{ref6}, BRISQUE\cite{ref35}, and NIQE\cite{ref34} are popular VQA models, used in streaming industry workflows, all of which quantify video quality (i.e. visual distortion severity) using models of human visual distortion perception. Streaming platforms currently employ a variety of evolving methods to assess various aspects of QoE, e.g., by blending Quality of Service (QoS) measurements of latency with VQA measures. QoE models like the Streaming Quality Index (SQI)\cite{ref10} and TV-QoE\cite{ref18} further enrich this assessment by factoring in rebuffering disruptions. Given QoE's intricate nature, many aspects of which are not currently addressable by any known neurophysiological models, most competitive QoE models now harness space-time machine learning, e.g. Video ATLAS\cite{ref14}, D-DASH\cite{ref19} and ITU-T Rec P1203\cite{ref37, ref38}, leveraging large datasets to create QoE algorithms that can be used to perceptually refine adaptive bitrate algorithms and enhance user experience.

Given the significant impact of network conditions on QoE, there has been increased emphasis on deploying network performance measurements to predict QoE. The rationale is that network performance fluctuations often correlate with degraded QoE, making them reasonably reliable indicators. Furthermore, from the ISP's point of view, this may be the only accessible data from which to predict video QoE. Recent methods improve on the use of simple network statistics\cite{ref20,ref21,ref22,ref23,ref24,ref25,ref26}, by applying machine learning methods to map network data to human judgments of QoE. QoE models like eMIMIC\cite{ref20} and BUFFEST\cite{ref21} statistically analyze network parameters, such as packet traces and HTTP requests, to gauge user QoE.

\section{CONSTRUCTION OF THE LIVE-VIASAT SATELLITE QoE DATABASE}
\label{sec3}


In the past, simulations of network-induced temporal distortion patterns have dominated the landscape of Quality of Experience (QoE) databases. These simulations have attempted to model real-world network congestion, manual bandwidth adjustments, and stall patterns. However, the limitations of these efforts lie in the fact that simulated distortions may not adequately capture real world situations, and generally poorly represent authentic distortions that occur in practice \cite{7327186, 7932975}. With these considerations in view, we made the following LIVE-Viasat Real-World Satellite QoE Database.

\subsection{The Viasat Satellite Dish Installation}
Since we wanted to capture videos under conditions closely resembling real-world video content and impairment scenarios, we deployed an authentic satellite network provided by Viasat, Inc. Viasat is recognized for its significant contributions and advancements in the domain of satellite video communications. Their communication satellites have adequate capacity to offer extensive coverage and many channels of high-speed internet capable of carrying video services.

\begin{figure}[!t]
\centering
\includegraphics[width=0.25\textwidth]{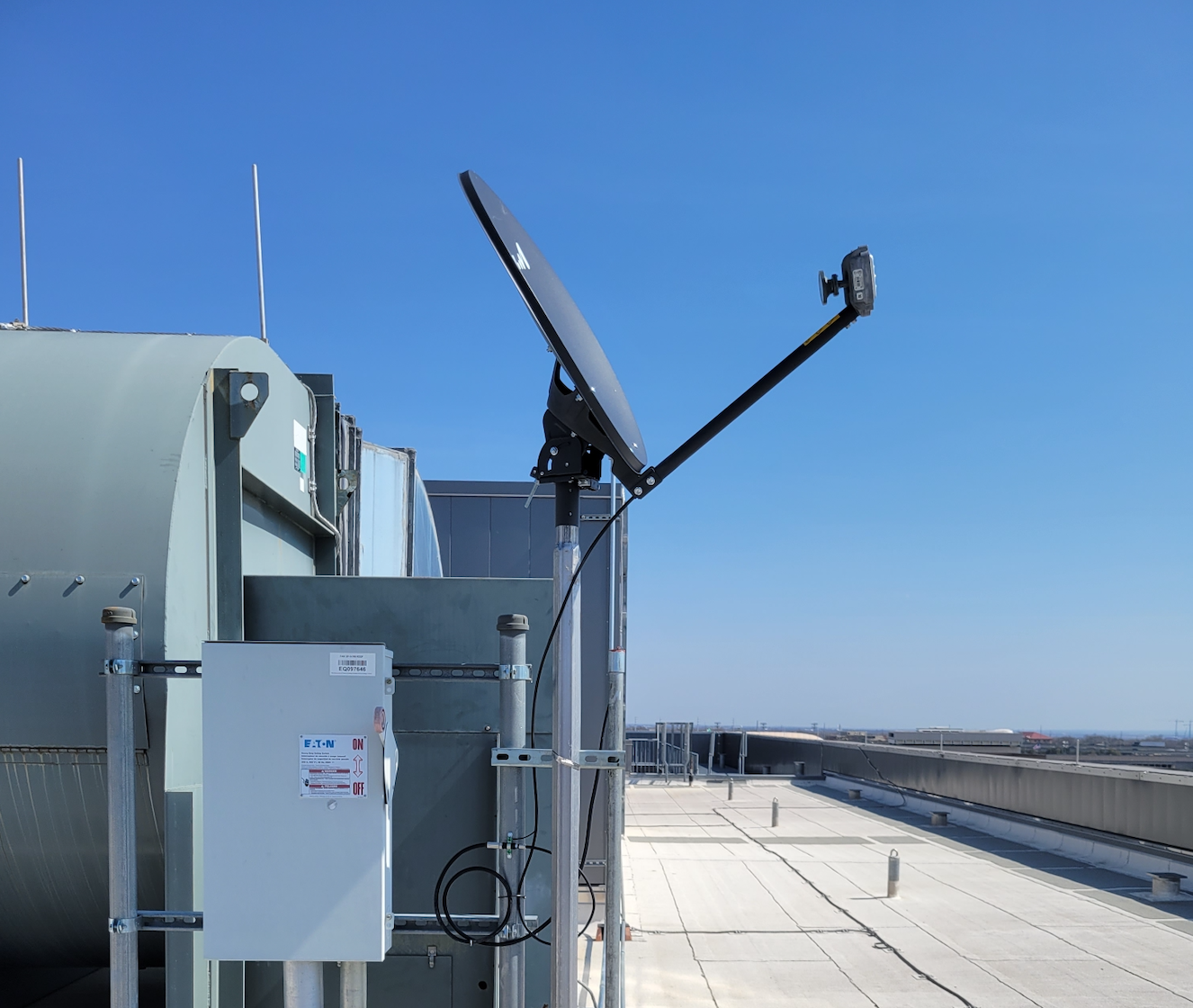}
\caption{Placement of the Viasat satellite dish atop the Engineering Education and Research (EER) building at the University of Texas at Austin (UT-Austin).}
\label{fig1}
\end{figure}

As shown in Fig. \ref{fig1}, Viasat installed a consumer-grade satellite dish atop the building housing our laboratory. The correct positioning, pointing, and overall configuration of the dish were executed by Viasat engineers, ensuring the effective capture of the satellite video signals. To optimize the process of data collection, UT-Austin technical support connected the dish output via high-speed Ethernet to a dedicated desktop research workstation in the Laboratory for Image and Video Engineering (LIVE). Ensuring the preservation of the accuracy and reliability of the video data during capture, transmission from the receiver, and display constituted a fundamental principle underlying our data collection process. We isolated the capture workstation from all other networks, guaranteeing that all the data and impairments originated from the Viasat satellite network.

\subsection{Source Videos}

\begin{figure}[!t]
\centering
\includegraphics[width=0.3\textwidth]{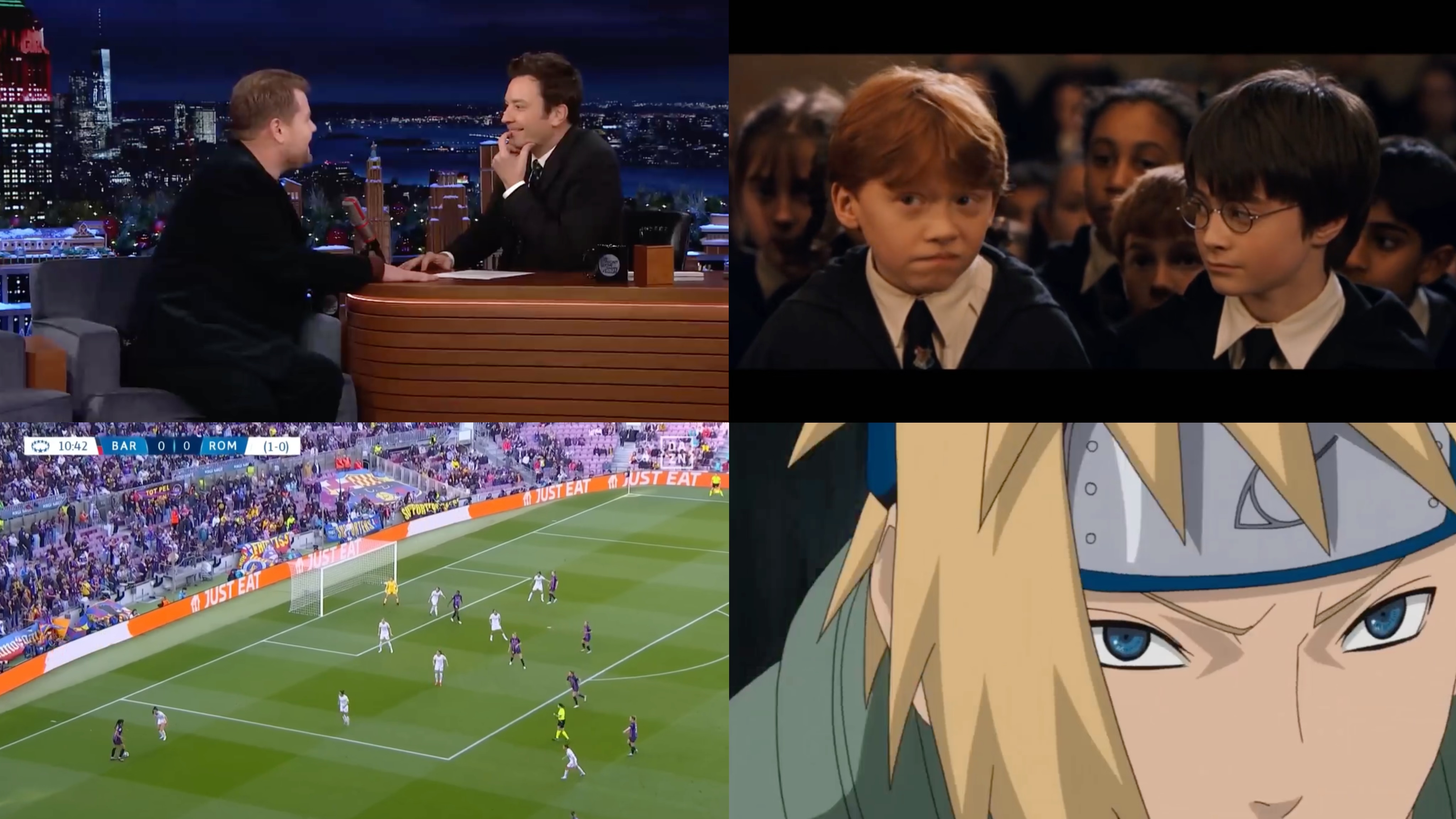}
\caption{A few frames from the LIVE-Viasat Real-World Satellite QoE Database.}
\label{fig2}
\end{figure}

We curated a collection of 27 user-generated High Definition videos sourced directly from the YouTube platform, which offers a diverse range of publicly accessible streaming video sequences across several content categories, including cartoons, concerts, video games, movies, news, sports, and talk shows. The video contents were classified into seven broad groups, as depicted in Table \ref{tab:table1}, presenting a wide range of attributes, thereby offering a more diverse assortment of video genres and qualities than those prevalent on most streaming platforms. The videos were captured by direct and exact screen capture. Several frames from these publicly available video sequences are depicted in Fig. \ref{fig2}. 

\begin{table}[!t]
\caption{Numbers of videos in each of six content categories. \label{tab:table1}}
\centering
\begin{tabular}{rr}
\toprule
\textbf{Category} & \textbf{Number of Videos} \\
\midrule
Sports & 9 \\
Concert & 3 \\
Game Video & 1 \\
Movie & 9 \\
Talkshow \& News & 4 \\
Vlog & 1 \\
\bottomrule
\end{tabular}
\end{table}

\begin{figure}[!t]
\centering
\includegraphics[width=0.3\textwidth]{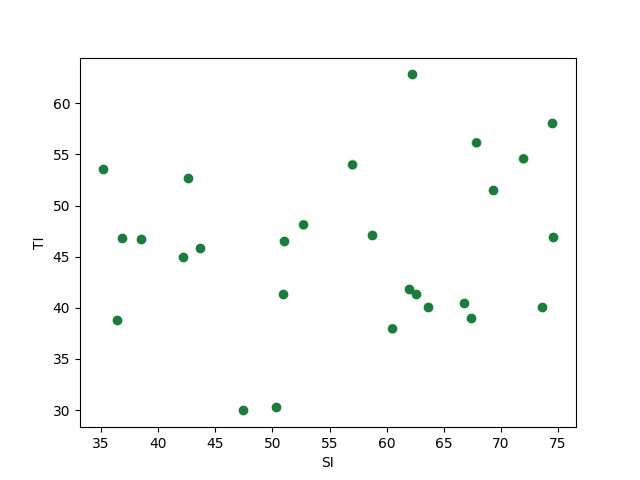}
\caption{Spatial Information (SI) plotted against Temporal Information (TI) of the 27 YouTube source videos.}
\label{siti}
\end{figure}

The streaming video in question, collected from YouTube, employs adaptive bitrate (ABR) streaming, a technique optimized to enhance the viewing experience by dynamically adjusting video quality based on network conditions. This video is segmented, with each segment displaying variations in bitrate, resolution, and frames per second (fps), highlighting the ABR technique's ability to balance video quality with smooth playback. Utilizing the efficient and effective AV1 codec, this streaming video demonstrates the adaptive capabilities of YouTube, which plays a pivotal role in determining the dynamic nature of the captured videos. The network protocol continuously adapts to varying conditions, resulting in fluctuating resolutions and frame rates of the videos. These adjustments depend on the strength of the network signal and the decisions made by adaptive streaming algorithms. The video was played on a display with a size of 1920x1080, typical for desktop streaming.

As is commonly done in experiments like this, we analyzed the amount of spatial and temporal information present in the videos. As noted in\cite{Chakravarty1995MethodologyFT}, Spatial information (SI) and temporal information (TI) are defined in this context as simple measures of spatial and temporal variations. Fig. \ref{siti} shows that the reference video content included in the LIVE-Viasat Real-World Satellite QoE Database contains a broad range of spatial and temporal variations.

\subsection{Video Data Collection Methodology}

We utilized the \emph{Surfmeter Automator} software, a commercial solution developed by AVEQ, to conduct automated measurements of video streaming sessions, including the capability to collect QoE-related data, perform high-quality video recordings (without audio), and gather satellite network packets. To prevent audio from influencing users' opinions of video quality, we did not record audio during the video recording process. The Surfmeter platform incorporates three essential components: 1) the \texttt{tcpdump} utility, for capturing network packets, 2) the YouTube 'Stats for Nerds' API in conjunction with the HTML5 API, for precise measurement of QoE parameters during video streaming, and 3) an implementation of an adaptive streaming QoE model (ITU-T Rec. P.1203, see Section~\ref{ssec:existing-qoe-models}).

\texttt{tcpdump} is commonly used for network diagnostics and capturing of data packets. It is a packet analyzer that operates in an open-source framework, providing users with a way to easily capture and dump packets traversing a network, thereby providing a snapshot of the network's activities. The YouTube 'Stats for Nerds' API offers a means to integrate YouTube videos into various applications and interact directly with the video player. By utilizing this API, developers and researchers can monitor various aspects such as stalls, bitrates, and rebuffering events.

Since we collected streaming videos played over satellite networks in the real world, we could not manually control the types or durations of rebuffering events, or variations in resolution and bitrate. We repeatedly played each source video more than 100 times and manually selected as many videos as possible, hence we were able to capture and record videos afflicted by many different types and degrees of distortion. Rebuffering events were observed to often occur simultaneously with decreases in bitrate and resolution.

Our dataset consists of 179 video clips, each with a duration between 30 and 60 seconds, procured through two distinct network mediums. Specifically, 161 videos were captured using the Viasat satellite network under two operational configurations: low-priority and normal modes. The low-priority mode restricts user bandwidth, resulting in more frequent and severe distortion levels compared to view videos transmitted in normal mode. Under the most typical circumstances, Viasat users view videos transmitted in normal mode, with the low-priority mode only occurring in exceptional situations, such as when a user's data usage exceeds their allotted limit. These configurations are representative of real-world user scenarios and enable the documentation of a broad spectrum of video distortions in terms of frequency and severity. By adopting this dual-source approach, we successfully compiled a diverse collection of video streams that are variably affected by different distortion types and levels, allowing for a nuanced evaluation of network service quality. The remaining 18 videos were acquired via the \emph{Wired General Network} at The University of Texas at Austin. This selection of very high-quality, identically displayed videos was purposefully acquired to serve as a comparative benchmark, representing user experiences typical of mainstream fiber optic networks, versus satellite network services, like those provided by Viasat.

\section{SUBJECTIVE STUDY}
\subsection{Subjects and Study Set-up}

Due to the substantial volume of video recordings amassed, the process of acquiring comprehensive subjective judgments of each video clip was a complex undertaking. We conducted a strategic division of the 179 video clips into 6 collections, which have about 30 videos each. When curating these collections, we included a balanced mix of videos: a selection from the satellite network's low-priority mode, often of lower quality; a set from the satellite's normal mode; and a group of high-quality videos captured on the UT-Austin high-bandwidth fiber optic network. Our aim was to ensure that each subject had a viewing time of approximately 2 hours spread over three sessions. Each participant was asked to view three of the six collections in a sequential rotation. For example, if participant $j$ viewed collections 1-3, participant $j+1$ engaged with collections 2-4. Precautions were implemented to guarantee the absence of any duplicated content during each session. Each subject was assigned to view one collection during each session of a duration of approximately 40 minutes. Over a span of three weeks, 54 students at The University of Texas in Austin, comprising both undergraduate and postgraduate students, participated in the study.

Each participant was provided with a set of guidelines and instructions at the beginning of their sessions. The instructions consisted of two phases: a written component followed by a spoken component. This aimed to ensure that participants had a clear understanding of their tasks with emphasis on the importance of evaluating each video without being influenced by its genre or topic matter. Each subject was situated at an approximate distance of 2.5 to 3 feet from the display but was encouraged to be comfortable.

\subsection{Study Interface}
The study was conducted using three desktop computers/ displays, each equipped with a display of 1080p resolution. This choice of resolution was made to ensure consistency and high-quality frames matching Viasat content delivery. Two of the workstations are configured with AMD Ryzen Threadripper PRO 5975WX@3.60GHz processors and two GeForce RTX 3090 Ti Graphics Cards, while the other is equipped with an AMD Ryzen 7 5700U@1.6GHz with Radeon Graphics. Each video was loaded entirely into memory before being presented to ensure that there was no additional, test-induced latency experienced during video playback. All three PCs were equipped with HP VH240a 23.8-inch Full HD 1080p IPS LED Monitors.

We created a bespoke interactive graphical user interface (GUI) using the Psychopy framework in the Python programming language. In order to optimize delivery to the participants, the video materials were prepared in advance in MKV format to enable uninterrupted viewing. To enhance the efficiency of the video streaming, the playback rate was adjusted to 60 frames per second, aligning with the refresh rate of the monitor.

\subsection{Testing Methodology}

Before beginning their task, each participant verbally affirmed that they possessed normal or corrected vision. Participants who normally wore corrective eyeglasses were instructed to wear them throughout their tasks. Once settled, each participant was introduced to the method of subjective evaluation. They were instructed to assess their continuous and overall viewing experiences by considering all visual aspects displayed on the screen. They were also advised against including the entertainment value of the video material as part of their quality assessments.

In order to familiarize participants with the interface, two independent video examples were presented, each demonstrating distortions that may be encountered. These videos were not included in the actual subjective data-gathering phase. From the second session onwards, it was assumed that the participants had acquired sufficient familiarity with the rating system and interface, so that orientation videos were no longer shown.

\begin{figure}[!t]
\centering
\includegraphics[width=2in]{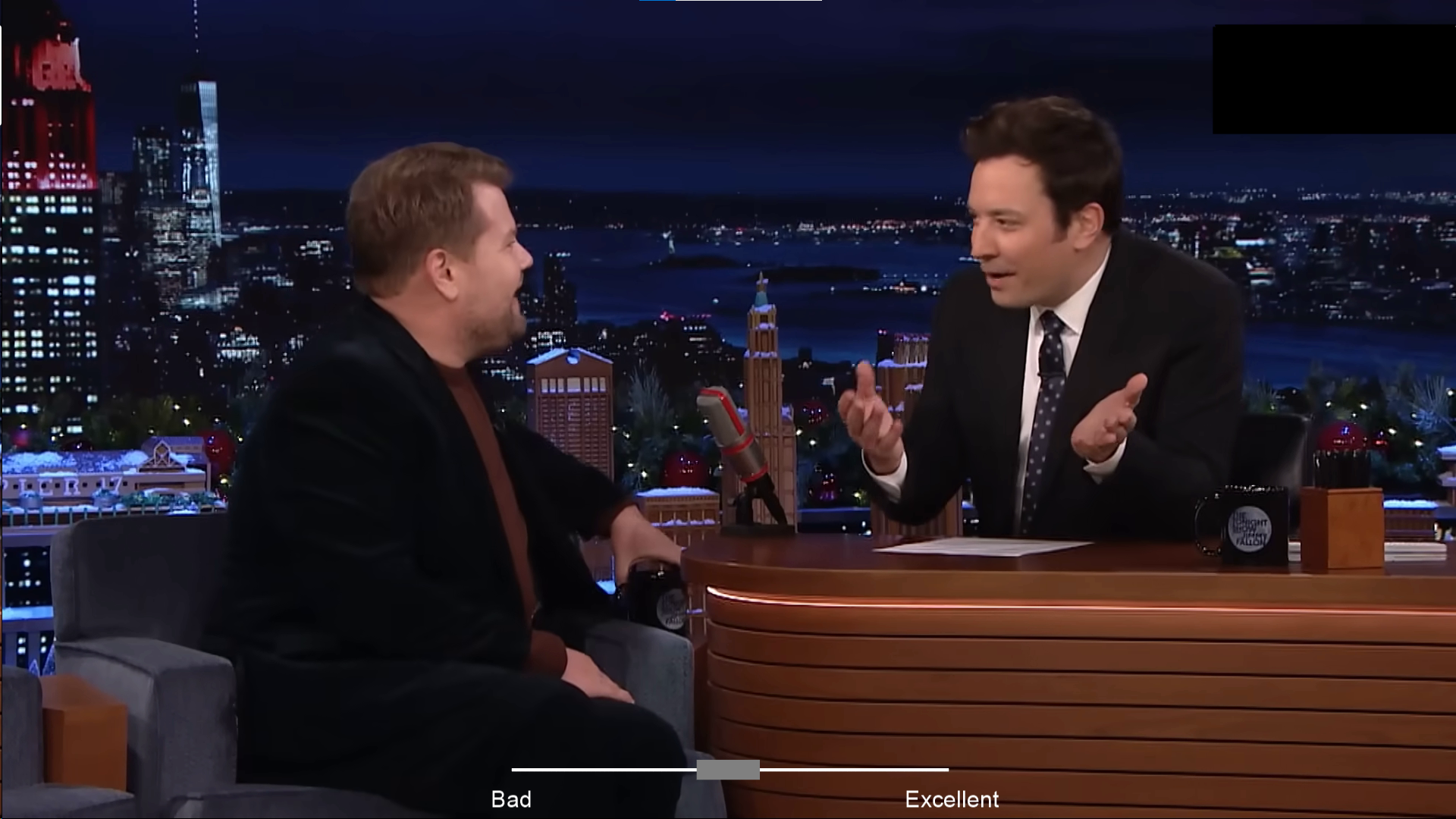}
\caption{Screenshot of the continuous rating bar (bottom) that continuously collects scores as a video is played.}
\label{fig3}
\end{figure}

\begin{figure}[!t]
\centering
\includegraphics[width=2in]{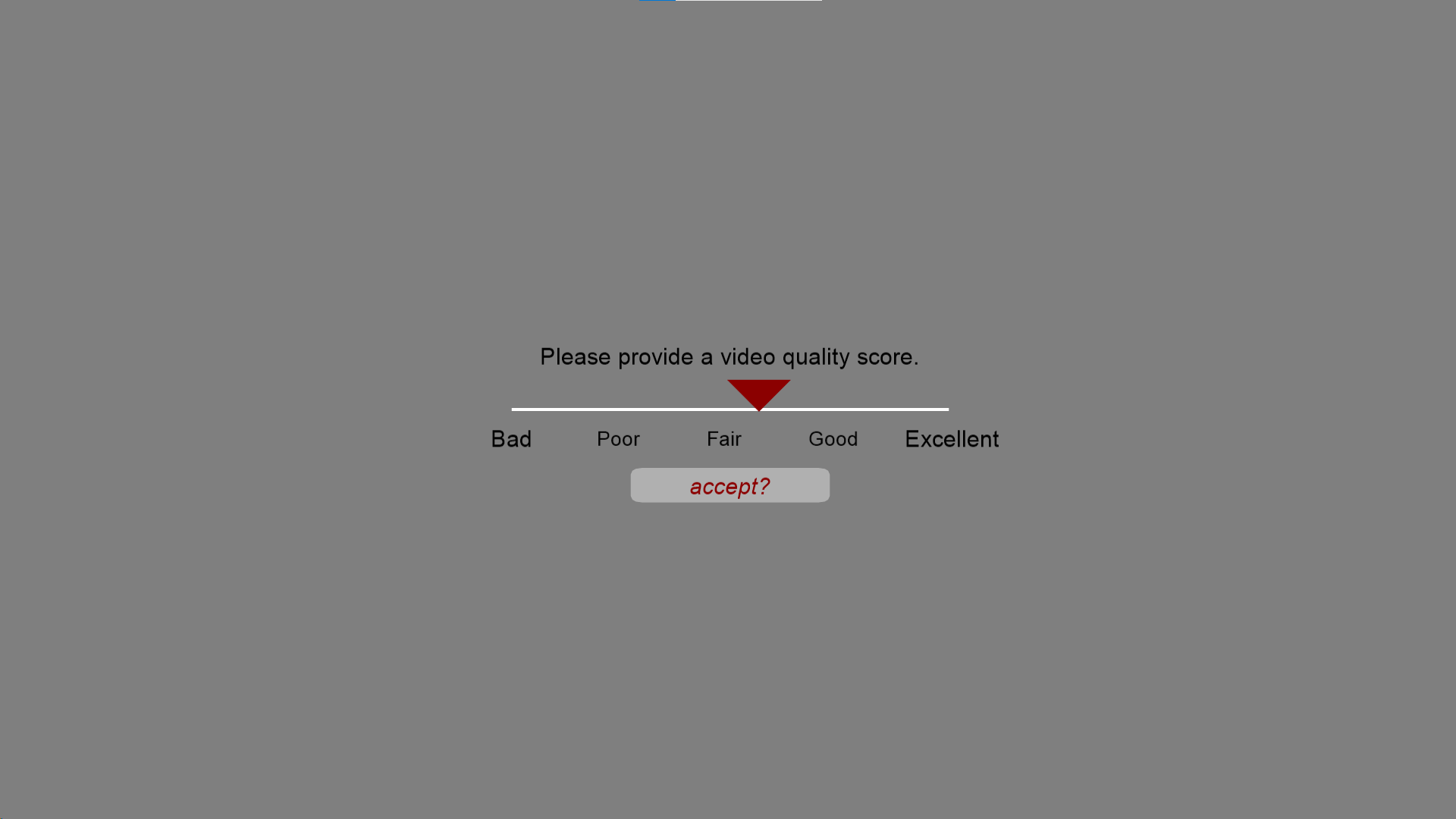}
\caption{Screenshot of the rating scale used to determine endpoint QoE once each video has finished playing.}
\label{fig4}
\end{figure}

In each session, the video clips were sequentially presented, accompanied by a continuous rating bar that remained visible at the bottom of the display during the video presentation as shown in Fig. \ref{fig3}, including Likert markings ranging from 'Bad' to 'Excellent', aligning with the ITU-R Absolute Category Rating (ACR) scale \cite{itu-t2}. Participants had the ability to manipulate their ratings using a computer mouse. Every second, 60 evaluations were captured, with each evaluation reflecting the real-time quality assessment of a specific frame. After each video played, participants were instructed to offer an overall rating of that video using the rating bar shown in Fig. \ref{fig4}.

\section{PROCESSING OF SUBJECTIVE SCORES}

\subsection{SUREAL Score Calculation}

Subjective ratings and scores, while highly valuable, inevitably contain noise and variability. To address this, a variety of standard methods exist for post-test screening. Among these, we opted for the SUREAL method \cite{ref28,ref29} due to its enhanced precision in uncovering bias and inconsistency issues, and its effective management of outliers through a "soft" rejection strategy.

The opinion scores $S_{ijr}$, are represented as stochastic variables having three components: 

\begin{equation}
    S_{ij} = \psi_{j} + \Delta_{i}+ v_{i}X. 
\end{equation}


The variable $\psi_{j}$ captures the inherent quality characteristic of video $j$, $\Delta_{i}$ represents the bias of subject $i$, $v_{i}$ quantifies the non-negative inconsistency of subject $i$, and $X$ follows a normal distribution $\mathcal{N}(0,1)$. The parameters $\psi_{j}$, $\Delta_{i}$, and $v_{i}$ were estimated by maximizing the log-likelihood of the scores using the Newton-Raphson method.

\subsection{Normalization of Continuous Subjective Scores}

We standardized the ratings into Z-scores \cite{ref30} to account for the inherent diversity in each participant's interpretation and application of the quality scale.
Let $s_{ijkl}$ denote the continuous QoE score assigned by participant $i$ to frame $l$ of video $j$ during session $k$. 
The mean and standard deviation of the scores assigned by this participant in the specified session were calculated as follows:

\begin{equation}
    \mu_{ik} = \frac{1}{\sum^{N_{ik}}_{j=1}}\sum^{N_{ik}}_{j=1}\sum^{N_{j}}_{l=1}s_{ijkl}
\end{equation}
and
\begin{equation}
    \sigma_{ik} = \sqrt{ \frac{1}{\sum^{N_{ik}}_{j=1}N_j-1}\sum^{N_{ik}}_{j=1}\sum^{N_{j}}_{l=1}(s_{ijkl}-\mu_{ik})^2}.
\end{equation}

Subsequently, the Z-score of each frame is derived as:

\begin{equation}
    z_{ijkl} = \frac{s_{ijkl}-\mu_{ik}}{\sigma_{ik}}
\end{equation}

Here $N_{ik}$ is the number of videos in session $k$ viewed by subject $i$, and $N_j$ is the number of frames in video $j$. 

\subsection{Subject Rejection Using Continuous Scores}

The perception of video quality is inherently subjective, involving delays or latencies in human reactions. During extended periods of quality assessments, participants may experience declines in attention or motivation, which may slow their responses further. Towards reducing response latency and subjective variations, we employed Dynamic Time Warping (DTW) \cite{ref31}, similar to \cite{ref12,ref13}, to better align the temporal subjective responses with the distortion occurrence. The DTW algorithm demonstrates exceptional proficiency in the task of synchronization of time series data. The DTW algorithm is a highly effective method of synchronizing time series data, which operates by adjusting the temporal alignment of sequences to optimize their similarity.


We applied an upper bound of 5 seconds on the adjustments made using DTW. We arrived at this figure by conducting tests using upper bounds ranging from 2 seconds to no upper bound. The 5-second cap was identified as the best observed limit for aligning sequences.

We also implemented protocols to identify outliers, by removing participants whose ratings lie well outside of normal behavior because of misunderstandings regarding the rating process, inattentiveness, or negative intent. To obtain a sense of the overall feedback trend across all subjects, we calculated an average QoE waveform of each video by aggregating the (warped) responses from all the participants. The mean waveform was subsequently used as a reference against which the individual responses were compared.

Subject rejection using the continuous scores proceeded as follows:
\begin{enumerate}
    \item An average waveform for the QoE was calculated on each video, computing the mean of the warped continuous-time waveforms recorded by all participants for each frame.
    \item We measured the disparity of the temporal scores supplied by each participant by comparing them against the average QoE waveform.
    \item By analyzing the variations in DTW values, we were able to detect individuals who provided inconsistent feedback.
\end{enumerate}

Subject rejection was carried out at the video level. When there was a significant difference between the warped difference between the recorded waveform of a participant and the average QoE waveform, it was deemed that their feedback did not correlate with the overall consensus, and were assumed to be an outlier. To implement this, we utilized the adjusted boxplot method specifically designed for skewed data \cite{ref32}. 

\begin{figure}[!t]
\centering
\subfloat[]{\includegraphics[width=0.25\textwidth]{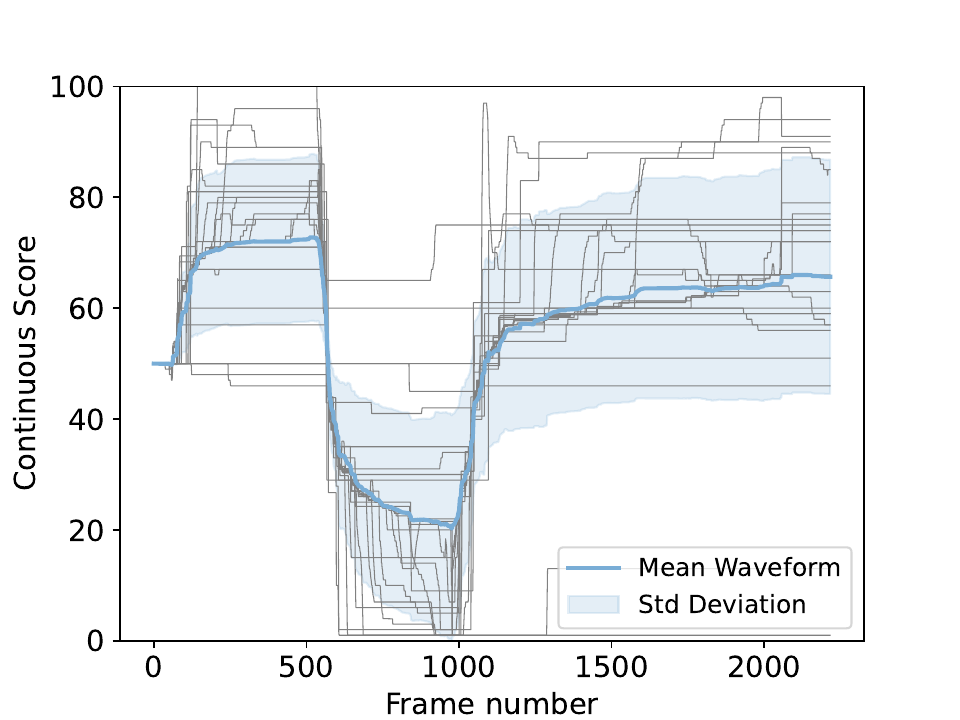}}
\subfloat[]{\includegraphics[width=0.25\textwidth]{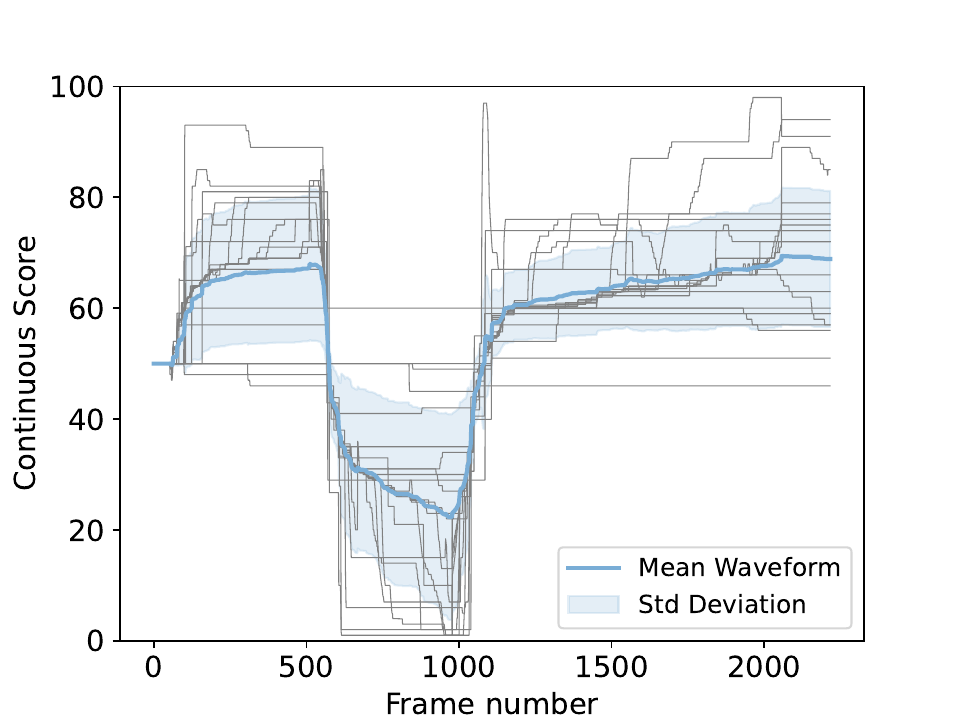}}
\caption{A pair of continuous (warped) QoE waveforms before and after subject rejection processing along with the average QoE waveform (blue line) and standard deviation around it (blue area). The left plot shows the original waveforms while the right plot shows the waveforms after subjective rejection.}
\label{waveform}
\end{figure}

Specifically, any person producing a warped difference relative to the average QoE waveform exceeding the value
\begin{equation}
    Q_3 + h_u(MC)IQR
\end{equation}
should be classified as an outlier. Here $Q_3$ is the third quartile, viz., 75\% of the data points that lie below Q3. $IQR$ is the interquartile range, the interval between the third quartile and the first quartile: $IQR = Q3-Q1$, representing the middle 50\% of the data and is a measure of data spread or variability. The factor $h_u(MC) = 1.5e^{bMC}$ provides a dynamic way to adjust outlier detection based on some measure $MC$. We assigned $b$ a value of $3$, as suggested in \cite{ref32}. The term $MC$ is the median-scaled difference between the two halves of a given data distribution. Fig. \ref{waveform} shows continuous QoE waveforms before and after subject rejection.



\section{ANALYSIS OF THE SUBJECTIVE DATA}
\label{sec6}
This study is intended to shed light on those factors that influence viewer behavior, such as stalling events, resolution, and bitrate, while also providing a tool for video engineers developing adaptive streaming algorithms that prioritize quality. Moreover, it is also intended to provide guidance to ISPs regarding the allocation of network resources and how to perceptually optimize end-user QoE. Here, we utilize continuous time and endpoint QoE scores to study the impact of objective QoE-related streaming characteristics, such as stall count, stall duration, stall location, bitrate, and resolution, as well as network parameters that are available to ISPs, such as data transferred size, throughput, and idle transmission time ratio, on the overall QoE.

\subsection{Effect of Stalls on QoE}

Stalls can substantially impact user experiences during the consumption of streaming videos. The authors of \cite{ref9, ref13} discussed the perceptual impact of simulated stalling events. The videos in the current study exhibit variations in the frequencies, positions, and durations of stalling events, which can significantly impact user Quality of Experience (QoE). An overview of further studies is provided in \cite{garcia2014quality}.

\subsubsection{Number of Stalls}

\begin{figure}[!t]
\centering
\includegraphics[width=0.3\textwidth]{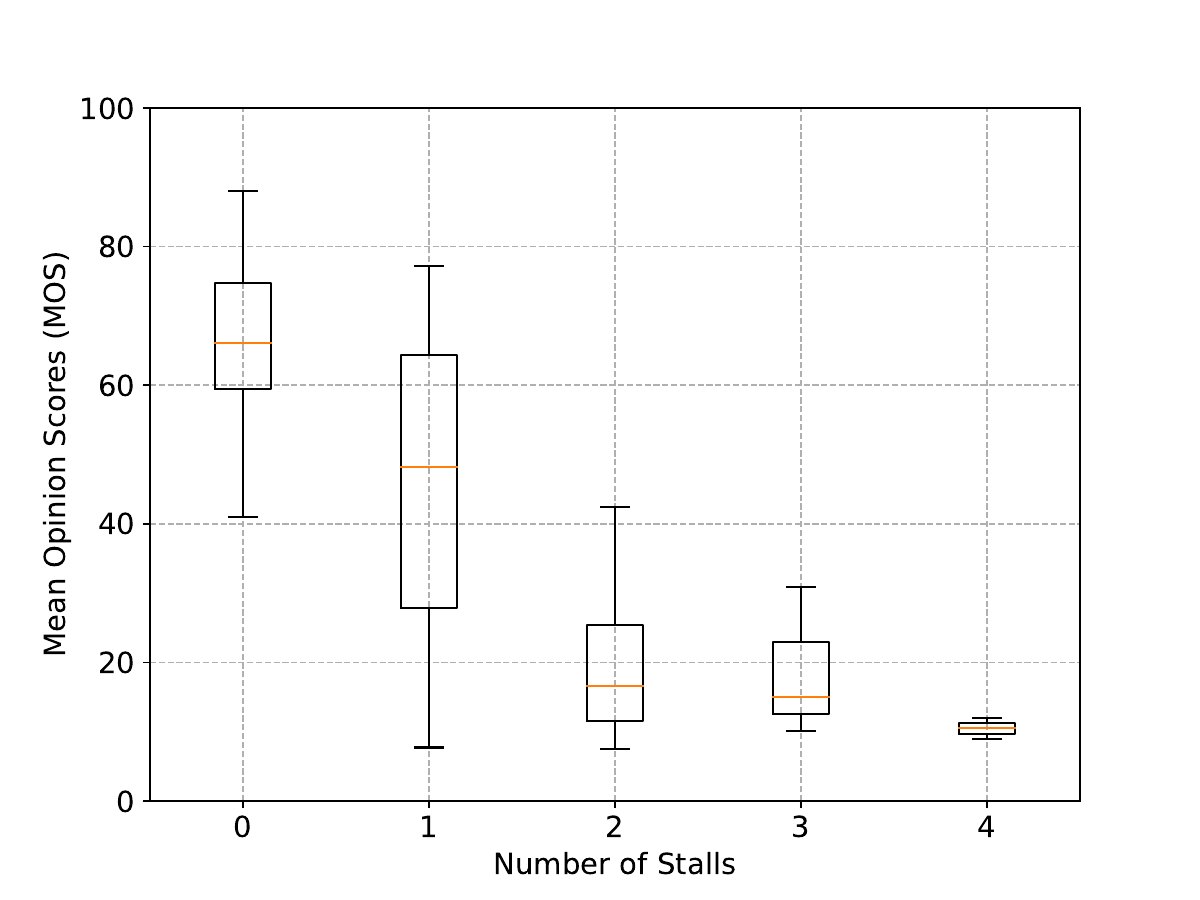}
\caption{Boxplot illustrating the distribution of endpoint QoE in relation to the number of stalls experienced during video playback.}
\label{stall num}
\end{figure}

We classified the videos based on the count of stalls that occurred, which varied from zero to four. As expected, we observed a negative correlation between the number of stalls and the endpoint QoE scores. As shown in Fig. \ref{stall num}, the median and interquartile range of endpoint QoE scores decreased with an increased number of stalls. This downward trend is indicative of the deterioration in user satisfaction as playback interruptions become more frequent. It is worth mentioning that on videos ranging from 30 to 60 seconds in duration, the decline in QoE was particularly pronounced when the number of stalls exceeded two. The QoE scores then plummeted below 20, indicative of a potential threshold of user tolerance. This suggests that viewers are relatively forgiving of up to two disruptions of short-duration videos but experience sharp declines in perceived quality with further interruptions.

\begin{figure}[!t]
\centering
\includegraphics[width=0.4\textwidth]{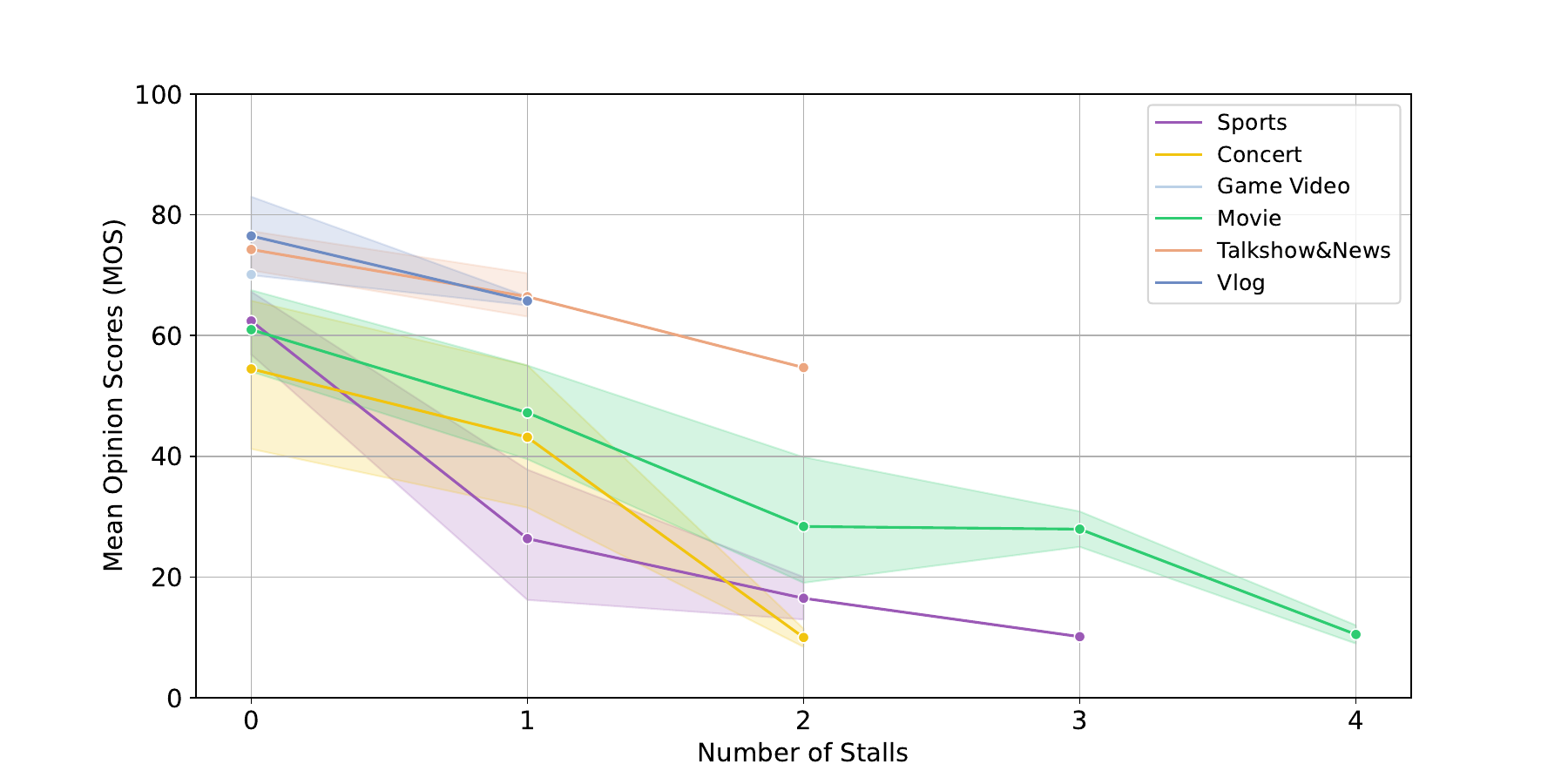}
\caption{Plot of endpoint QoE against the number of stalls for different video categories.}
\label{stall num_category}
\end{figure}

We also examined human responses against the number of stalls across video categories (Fig. \ref{stall num_category}). We found that the video category influenced the rate at which QoE decreases with stall frequency. Videos categorized under 'Sports' exhibited steeper decreases in QoE than videos in all the other categories. This is likely because sports content contains fast-moving action which may be disrupted or lost by stalls, causing a larger perceived reduction of QoE. 

\subsubsection{Duration of Stalls}
\begin{figure}[!t]
\centering
\includegraphics[width=0.3\textwidth]{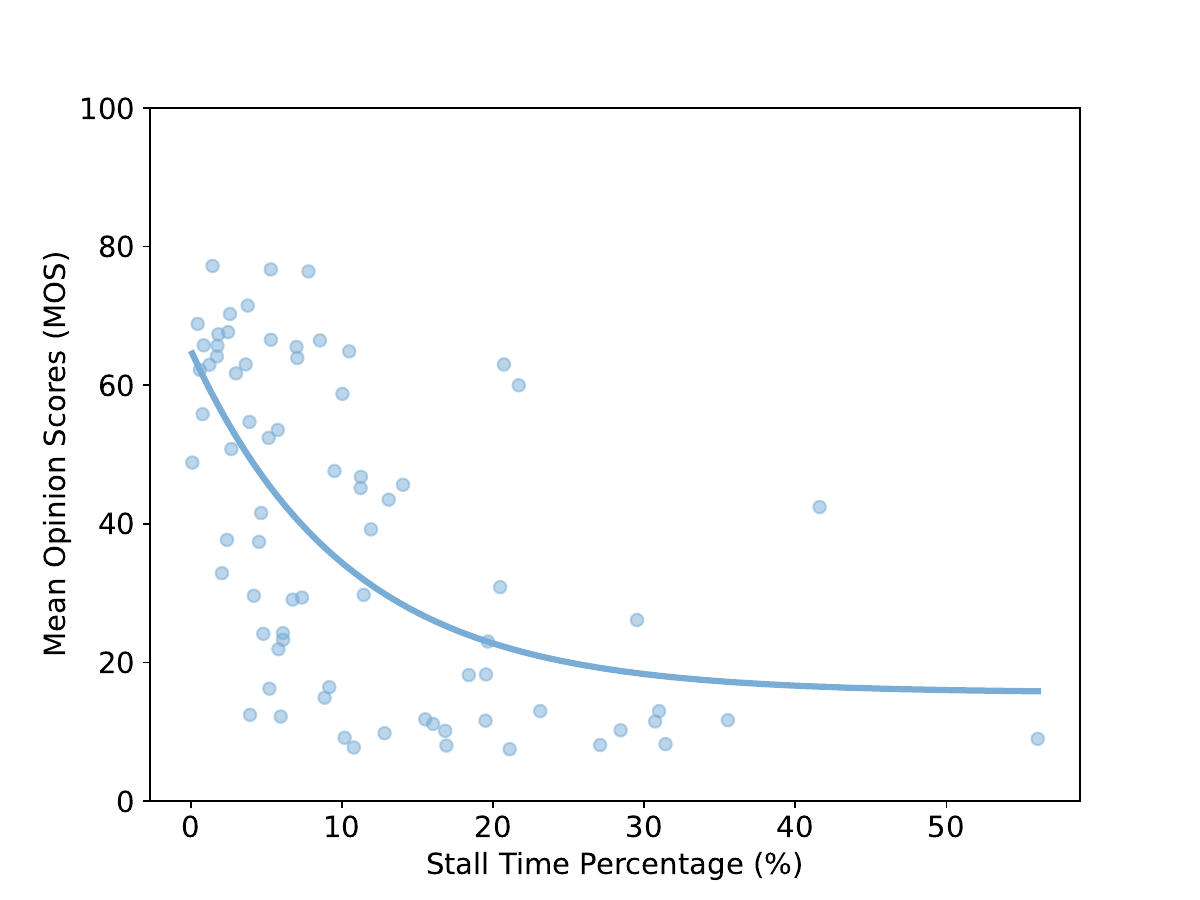}
\caption{Endpoint QoE scores plotted against the ratio of stall duration to video duration. A logistic fit to the data is overlaid.}
\label{stall duration}
\end{figure}

We computed the proportion of the overall duration of each video that was affected by stalls. We then conducted an analysis to examine the relationship between user MOS at the endpoint and the ratio of stall to video durations experienced during the playback of the videos, with the results shown in Fig. \ref{stall duration}. As the proportion of rebuffering time increased, there was a significant decrease in endpoint MOS.

\begin{figure}[!t]
\centering
\subfloat[]{\includegraphics[width=0.25\textwidth]{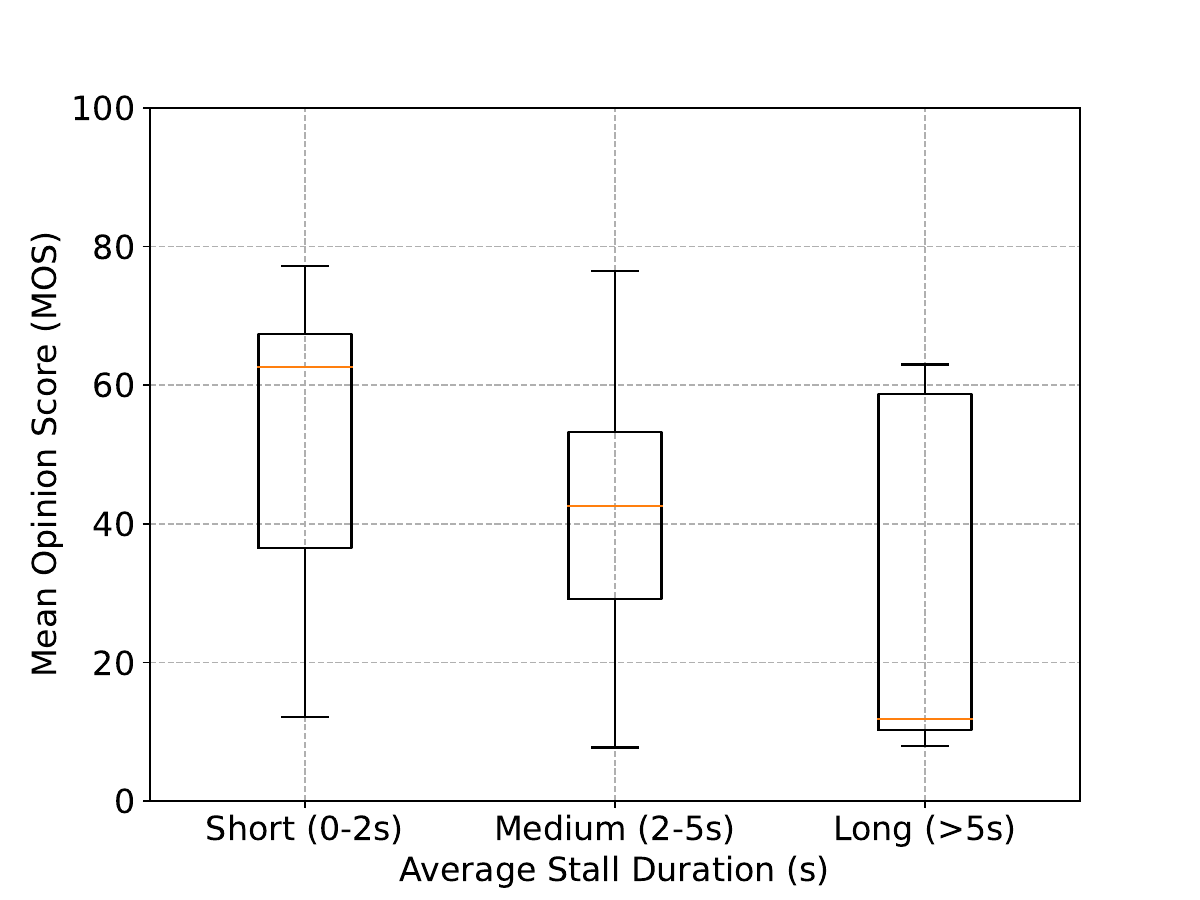}}
\subfloat[]{\includegraphics[width=0.25\textwidth]{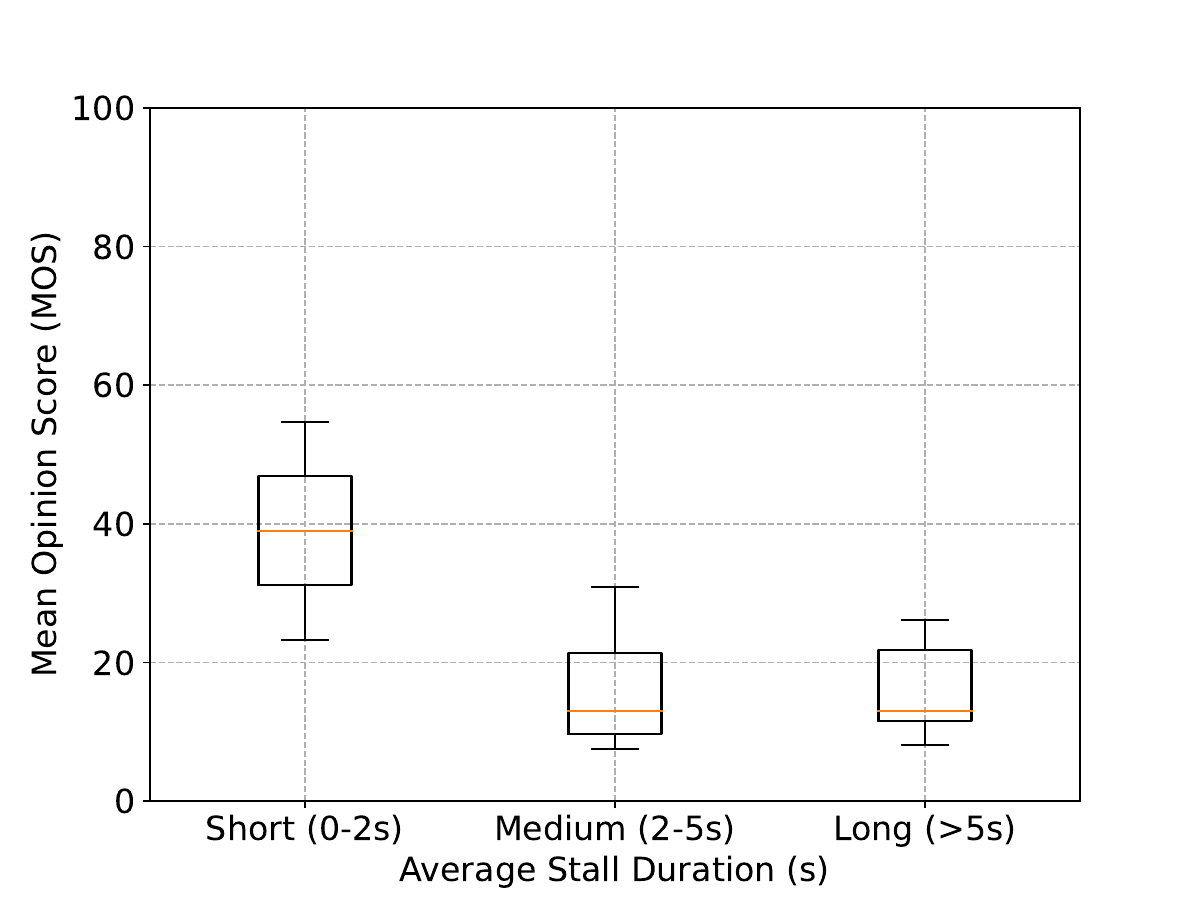}}
\caption{Boxplot analysis of MOS on (a) videos interrupted by a single stall; (b) videos containing two or more stalls. In both plots, the videos are categorized by average stall duration.}
\label{stall_duration2}
\end{figure}

Next, we analyzed the impact of stall duration on endpoint QoE by categorizing video playback interruptions into three segment durations: short (0-2s), medium (2-5s), and long ($>$5s). Fig. \ref{stall_duration2} (a) shows the endpoint QoE distribution across the stall durations of videos containing a single stall event. The observed declines in endpoint MOS from short to long stall durations indicate that as the stall durations increased, the spread of MOS scores widened and the median MOS decreased, signifying more pronounced dissatisfaction among viewers. Among videos containing multiple stalls, the human reactions to average stall durations showed a similar trend but bottomed out below 20 for even medium average stall durations as shown in Fig. \ref{stall_duration2} (b). This suggests a cumulative effect of interruption frequency on QoE, with the presence of multiple stalls having a greater impact on MOS than duration effects.


\begin{figure*}[!t]
\centering
\subfloat[]{\includegraphics[width=0.25\textwidth]{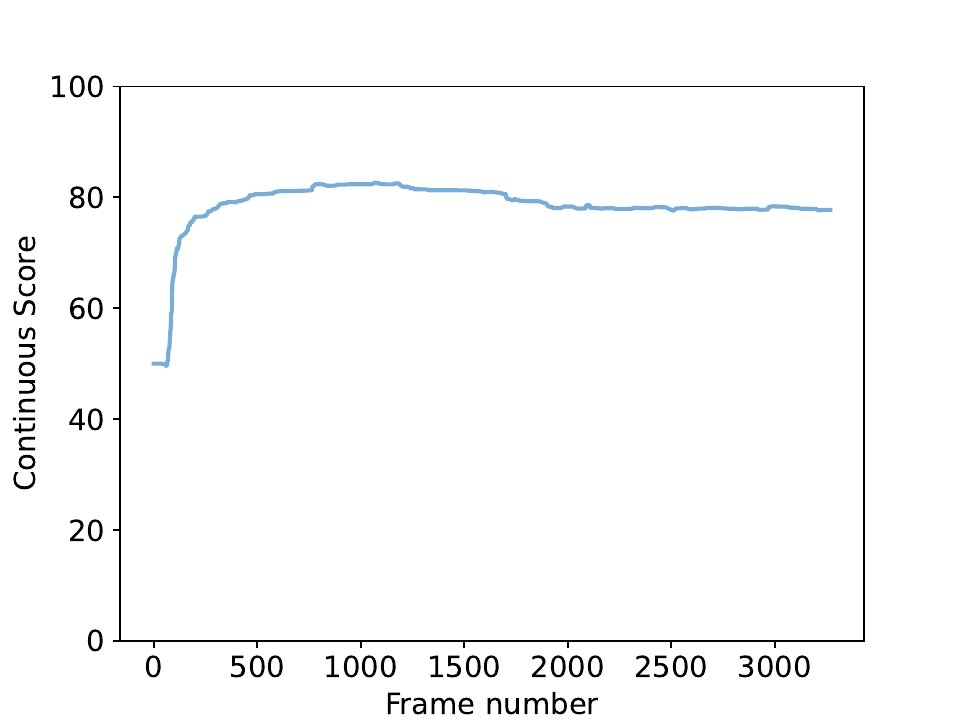}}
\subfloat[]{\includegraphics[width=0.25\textwidth]{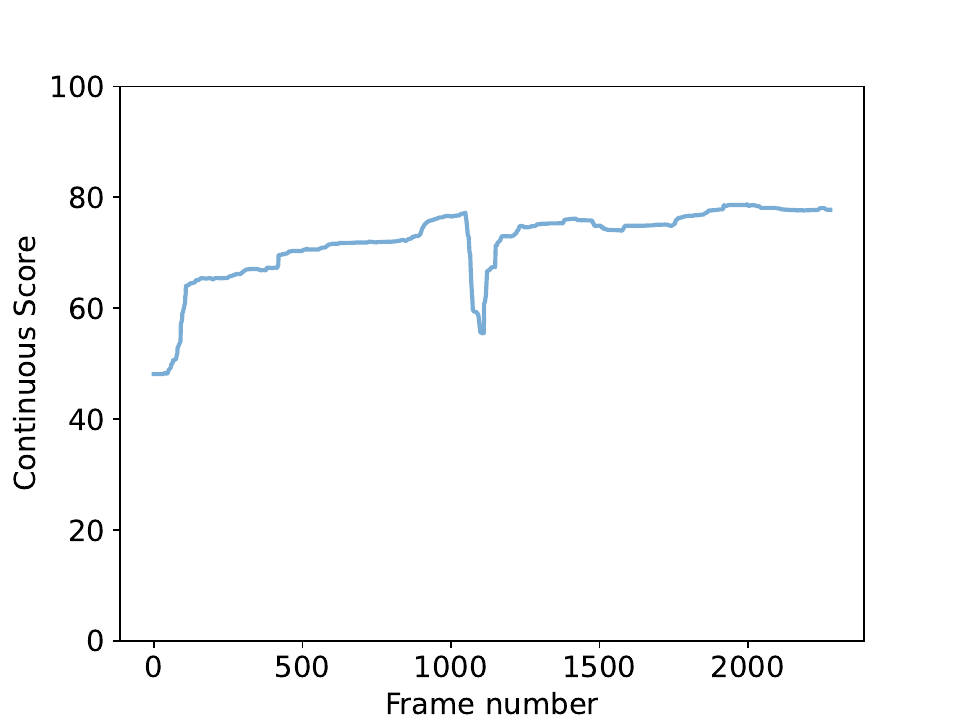}}
\subfloat[]{\includegraphics[width=0.25\textwidth]{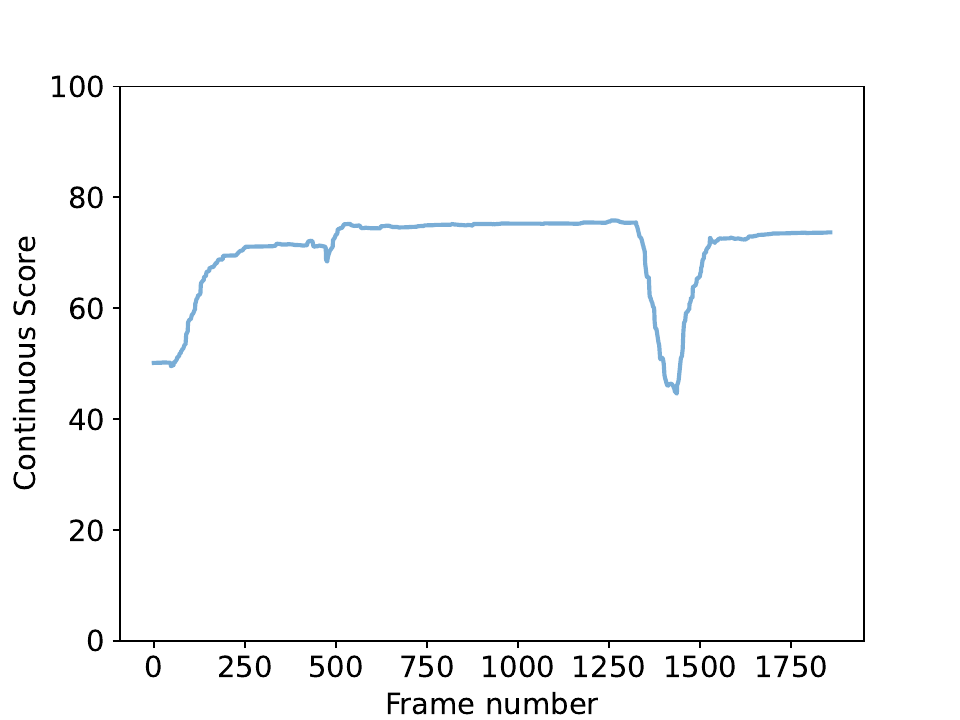}}
\subfloat[]{\includegraphics[width=0.25\textwidth]{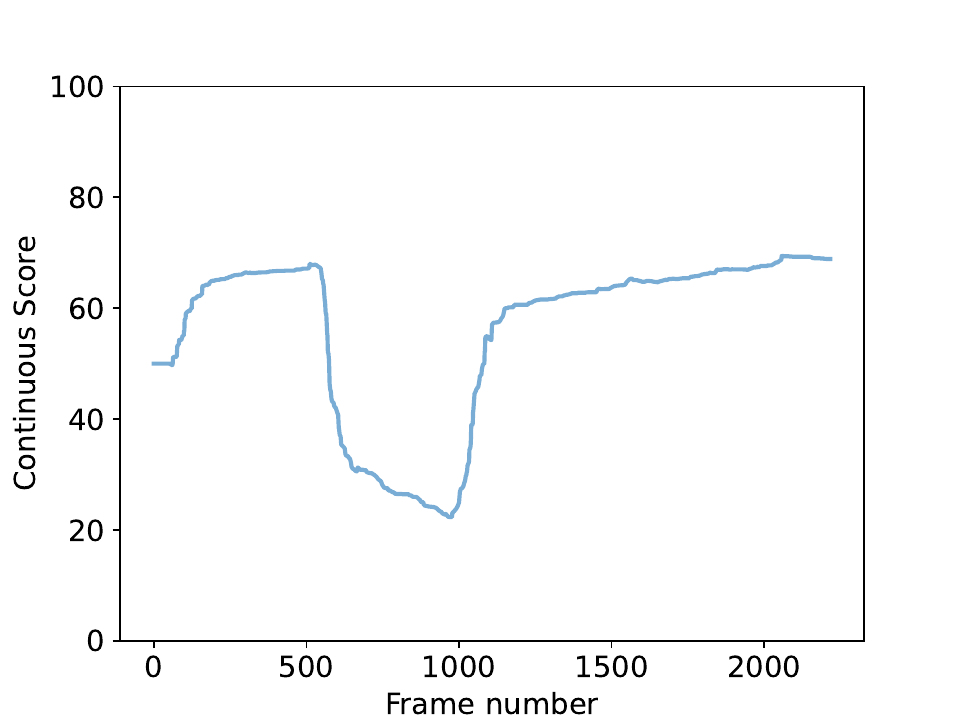}}
\caption{Continuous QoE scores of four video instances derived from the same source video (same spatial-temporal complexity). (a) No stall occurs. (b) A stall occurs at frame 1012 and lasts for 59 frames (about 0.98s). (c) A stall occurs at frame index 1293 of duration 131 frames (about 2.18s). (d) A stall occurs at frame index 511 of duration 460 frames (about 7.67s).}
\label{duration}
\end{figure*}

We also found interesting results regarding spatial and temporal complexity. If a stall occurred closer to the commencement of a low-complexity video and the spatial resolution was higher and did not change, then viewers tended to be more forgiving of the stall. Conversely, in videos characterized by higher spatial-temporal complexity and lower resolution, users were more sensitive to stall duration, especially when they occurred near the end. Among videos of similar temporal-spatial complexities, increases in the durations of individual stalls invariably led to a pronounced reduction in user satisfaction. Following a stall, the QoE tended to only partially recover, reflecting a hysteresis or memory effect \cite{ref33}. Fig. \ref{duration} plots instances of the continuous QoE scores of four video samples derived from the same source video. 

\subsubsection{Position of Stall}

When a video is interrupted by multiple stalls, it is of interest to examine how the positions of the first and the last stall influence users' overall QoE perception scores. 

\begin{figure}[!t]
\centering
\subfloat[]{\includegraphics[width=0.25\textwidth]{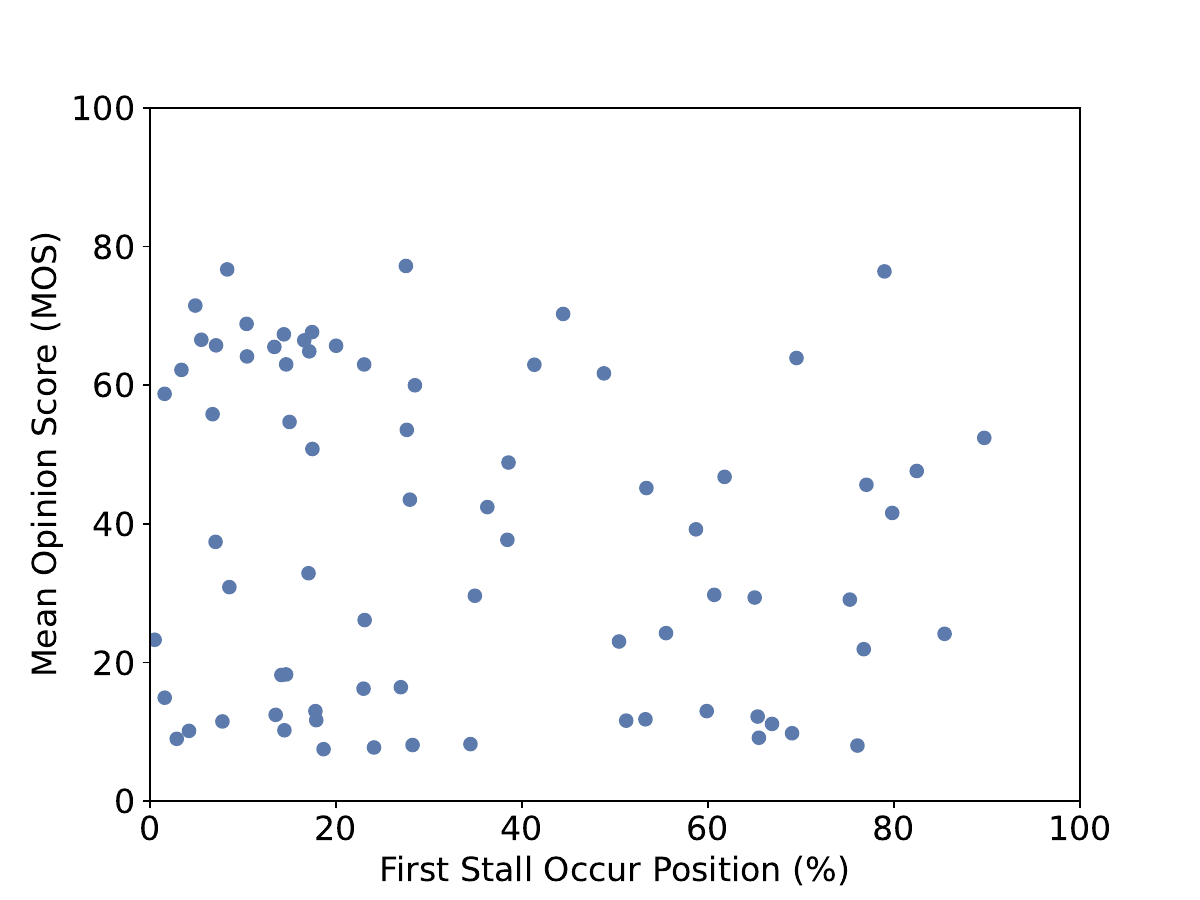}}
\subfloat[]{\includegraphics[width=0.25\textwidth]{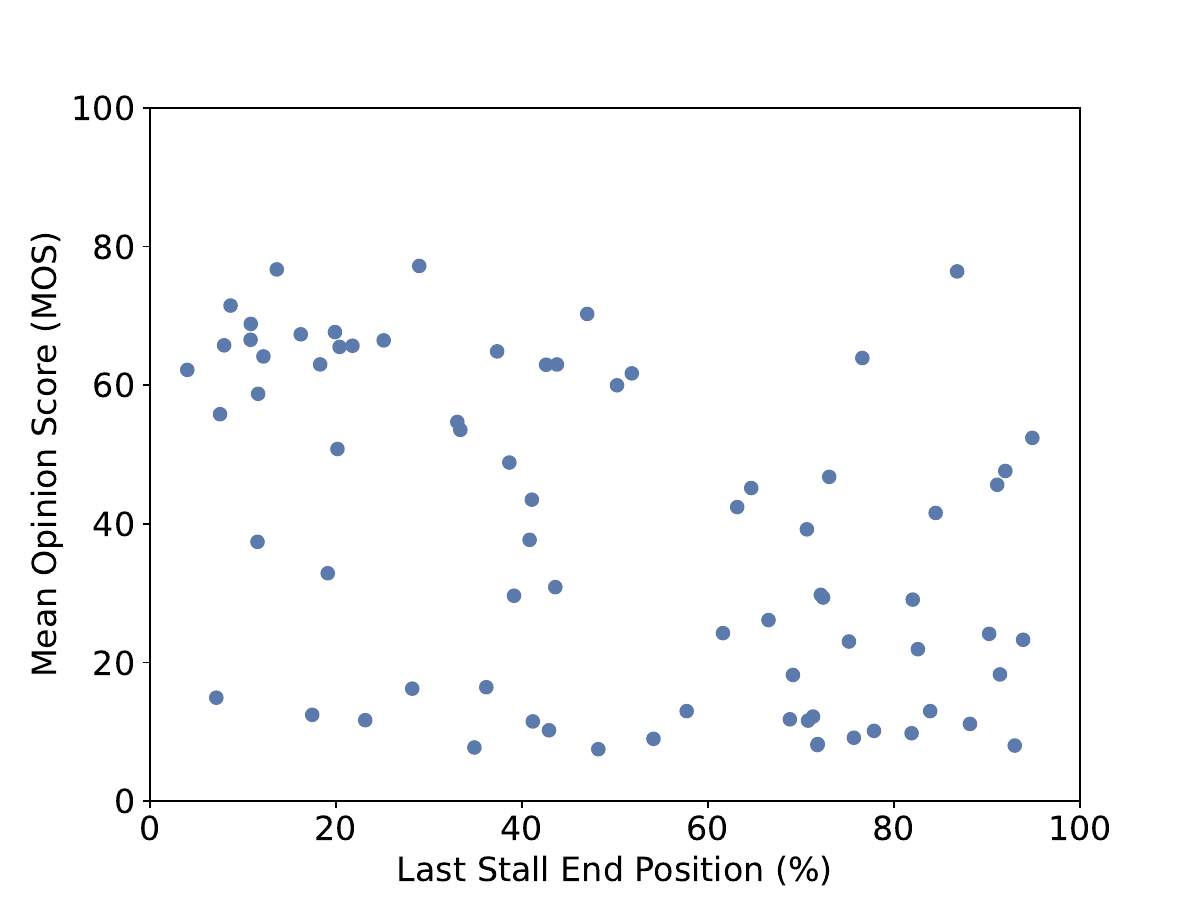}}
\caption{(a) Scatter plot of the position of each video's first stall against endpoint MOS. (b) Scatter plot of the position of the end of each video's last stall against endpoint MOS. In both plots, stall positions are expressed as a percentage of total duration.}
\label{stall_position2}
\end{figure}

Fig. \ref{stall_position2} (a) plots the positions of first stall occurrences, expressed as a percentage of total video duration, against the corresponding endpoint MOS. The data does not indicate a strong relationship; instead, there is a wide dispersion of endpoint MOS across all points of initial stall occurrence. This suggests that the impact of first stalls on QoE may be more heavily influenced by a combination of other factors.

The analysis of last stall events, as shown in Fig. \ref{stall_position2} (b), reveals a more pronounced trend. Stalls that conclude towards the latter part of the video are strongly associated with lower endpoint MOS values. This also is an expected recency effect\cite{weiss2014temporal}. While it is difficult to extrapolate recency effects observed on shorter videos to those on full-length productions ($>$30 min), it may be quite relevant when considering QoE of single shots or scenes.

\begin{figure*}[!t]
\centering
\subfloat[]{\includegraphics[width=0.25\textwidth]{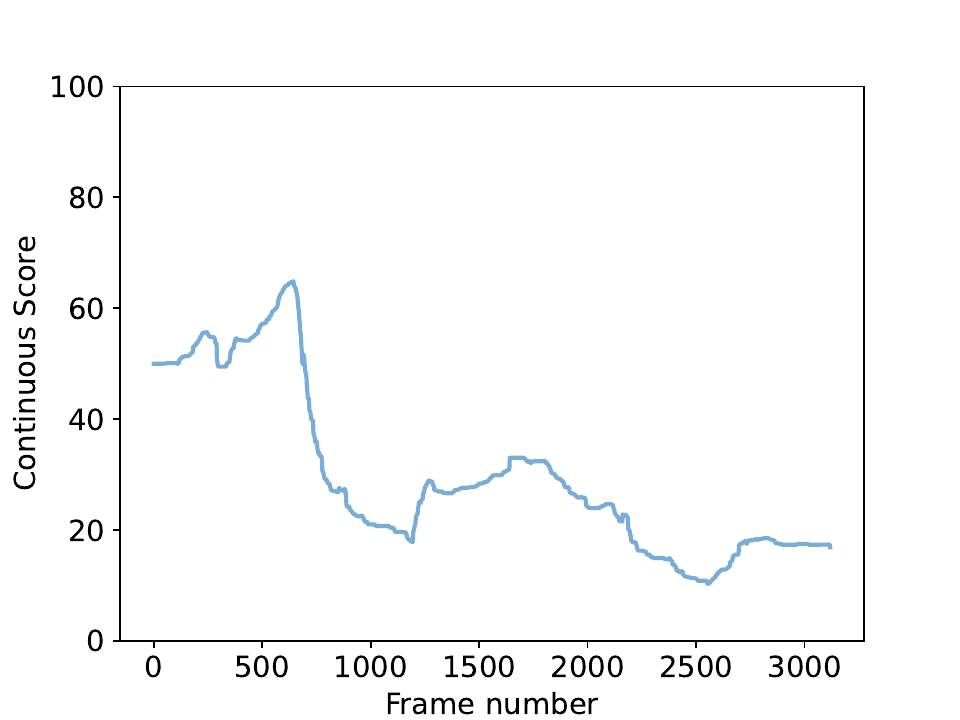}}
\subfloat[]{\includegraphics[width=0.25\textwidth]{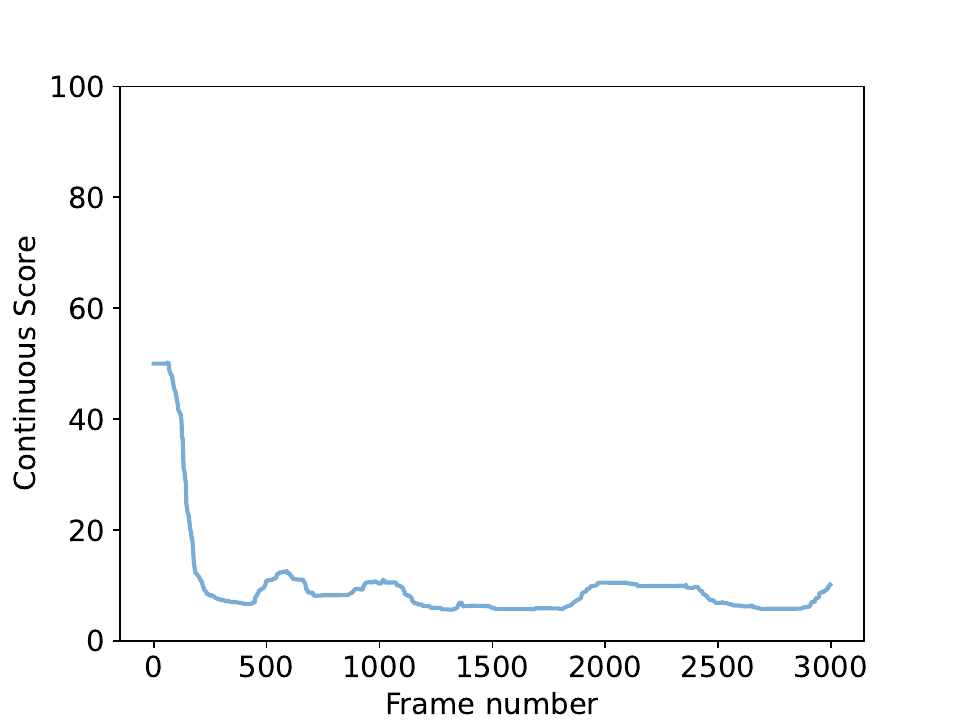}}
\subfloat[]{\includegraphics[width=0.25\textwidth]{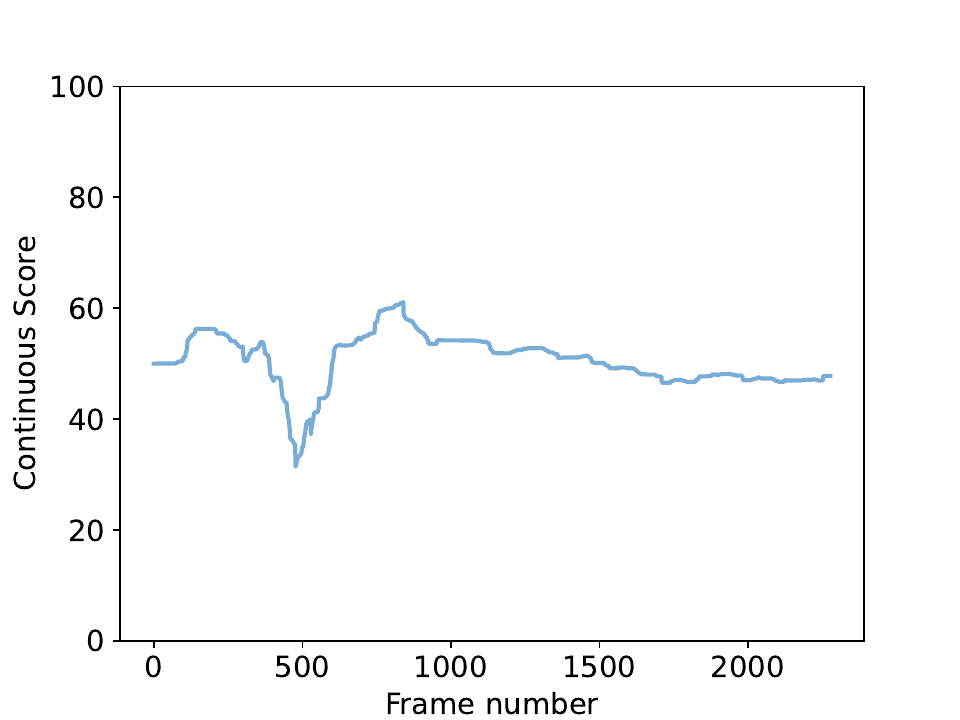}}
\subfloat[]{\includegraphics[width=0.25\textwidth]{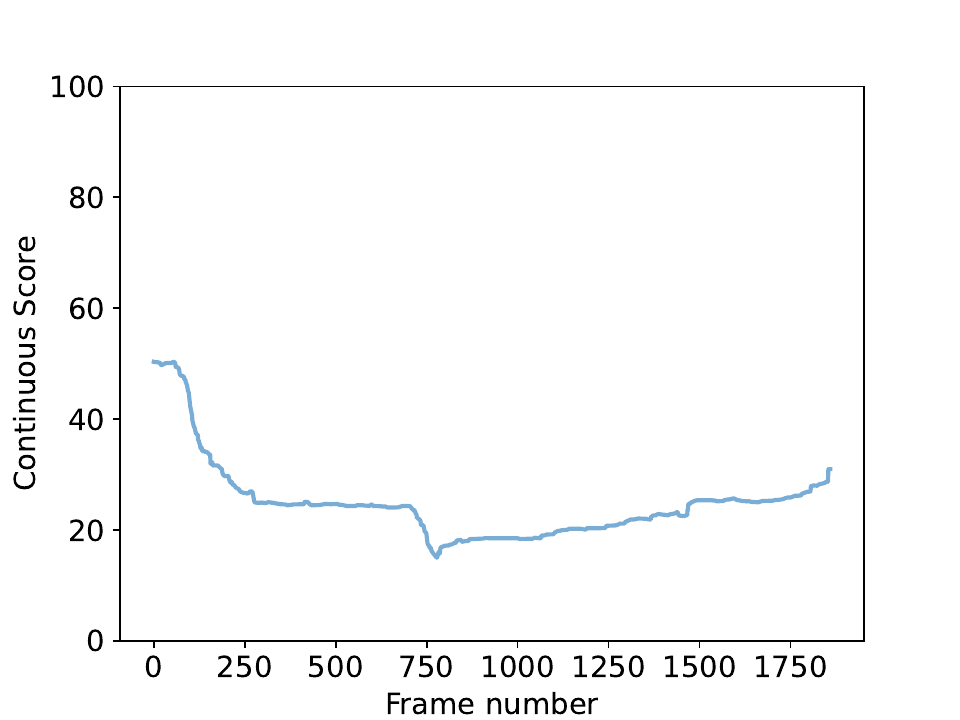}}
\caption{Continuous QoE scores of four video instances. (a). Two stalls occur, at frame indices 733  and 2064 of durations of 445 (about 7.42s) and 494 frames (about 8.23s) respectively. (b).  Four stalls occur at frame index 88, 553, 1250, and 2382 leading to a rapid QoE drop followed by relatively minor score fluctuations.  (c) and (d) are continuous QoE scores for two video instances of the same source content. (c). A stall of duration 61 frames (about 1.02s) occurs at frame index 398. (d). A stall of duration of 80 frames (about 1.33s) occurs at frame index 398.}
\label{position}
\end{figure*}

When stalls occur in succession, we observe a consistent drop in continuous MOS. This phenomenon is depicted in Fig. \ref{position} (a). However, it should be observed that in such instances, the video may be modified by increased compression or scaling, also affecting QoE. Under such circumstances, additional stalls may only induce minor score fluctuations, as illustrated in Fig. \ref{position} (b). Notably, the presence of multiple stalls frequently correlates with dips in resolution.

Furthermore, Figs. \ref{position} (c) and \ref{position} (d) highlight an important aspect of user satisfaction: when the MOS is high, indicating that the viewers were content with the video playback quality, any stalls caused pronounced declines in the real-time MOS. This decline was significantly sharper than when the continuous MOS had been at a lower level.

\subsection{Effect of Resolution on QoE}

\begin{figure*}[!t]
\centering
\subfloat[]{\includegraphics[width=0.25\textwidth]{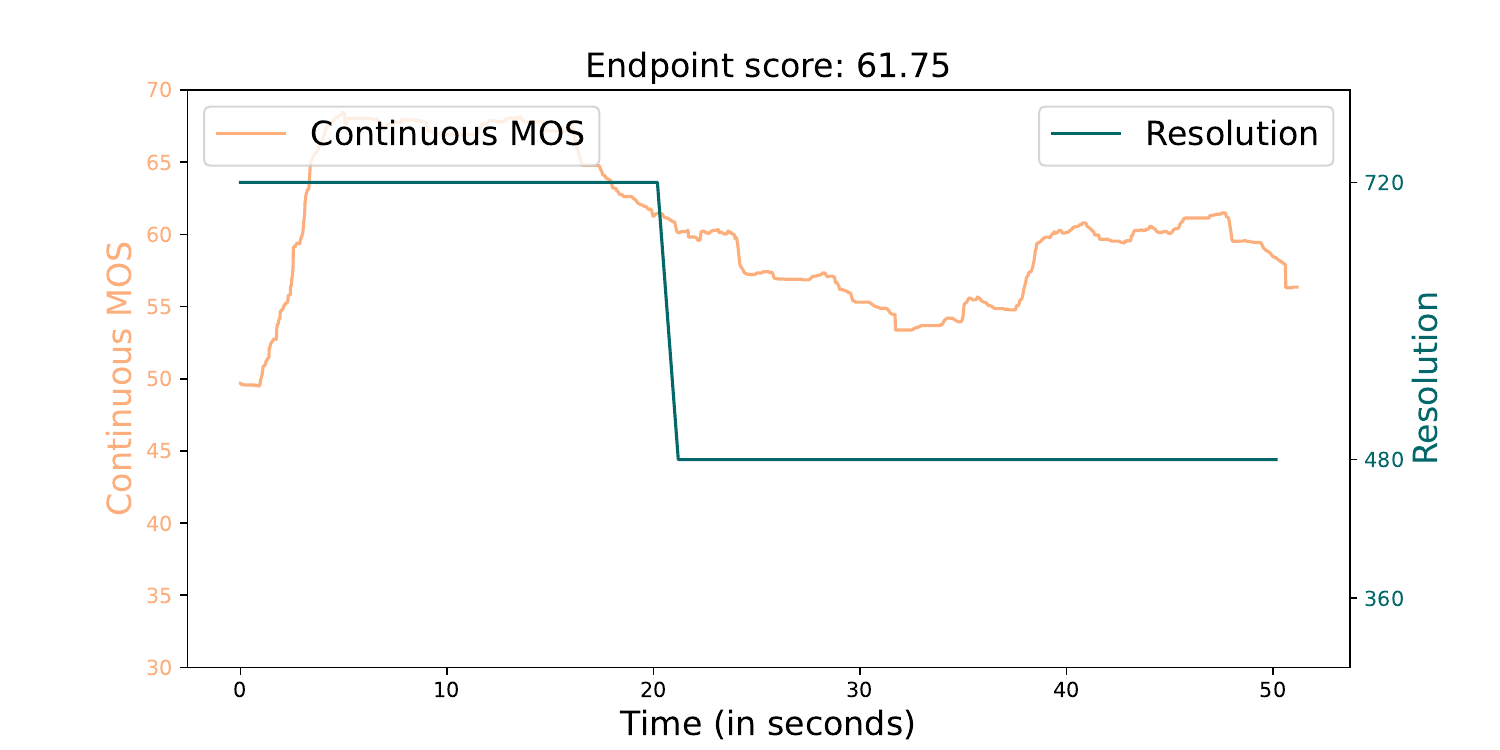}}
\subfloat[]{\includegraphics[width=0.25\textwidth]{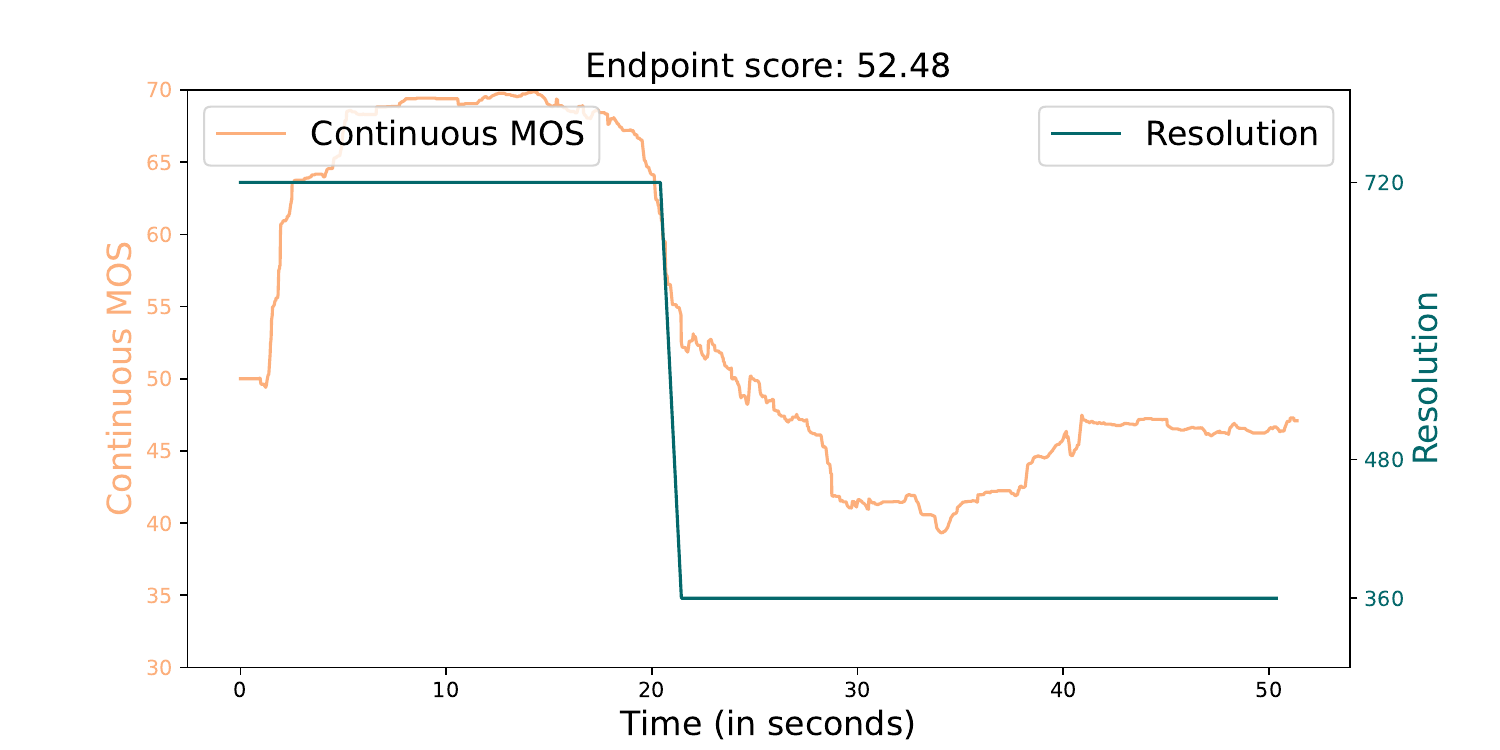}}
\subfloat[]{\includegraphics[width=0.25\textwidth]{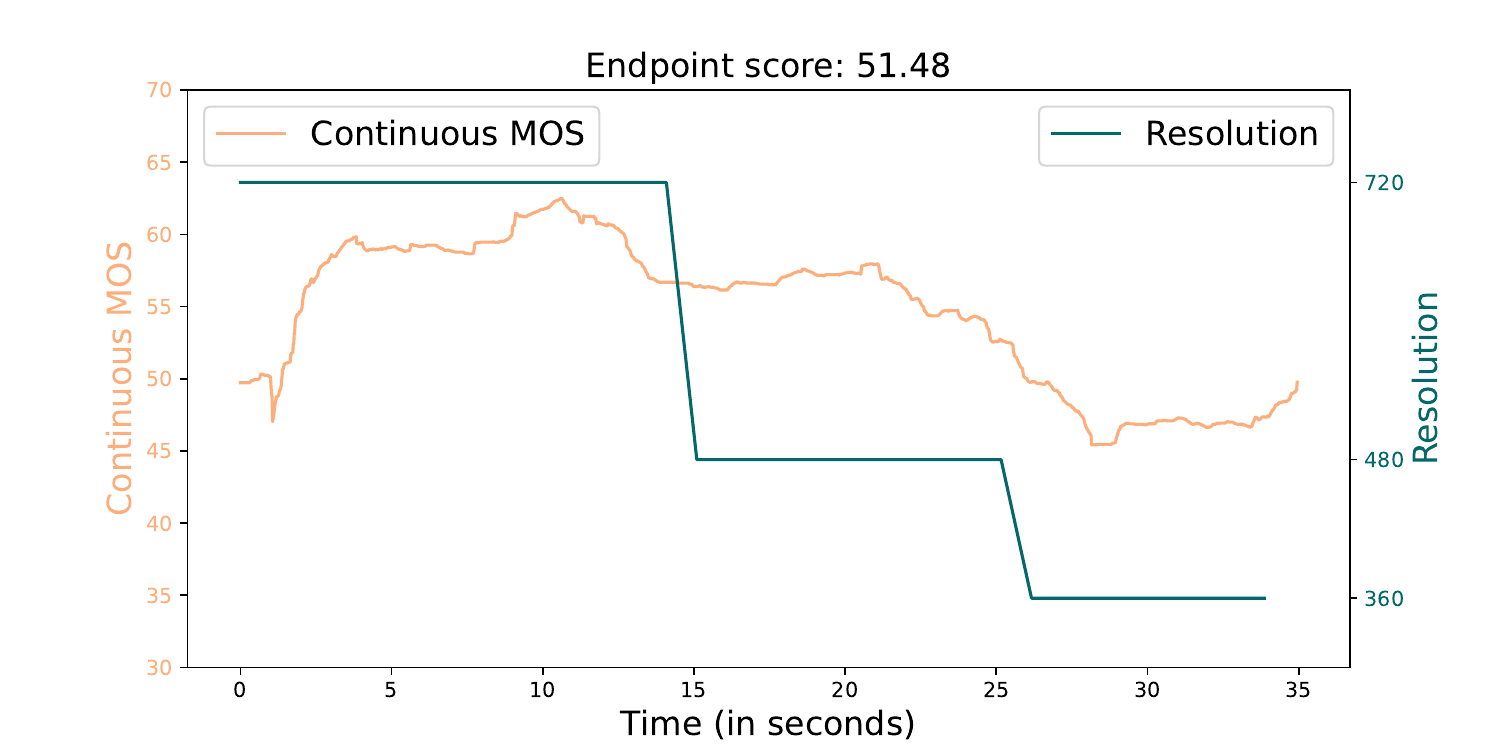}}
\subfloat[]{\includegraphics[width=0.25\textwidth]{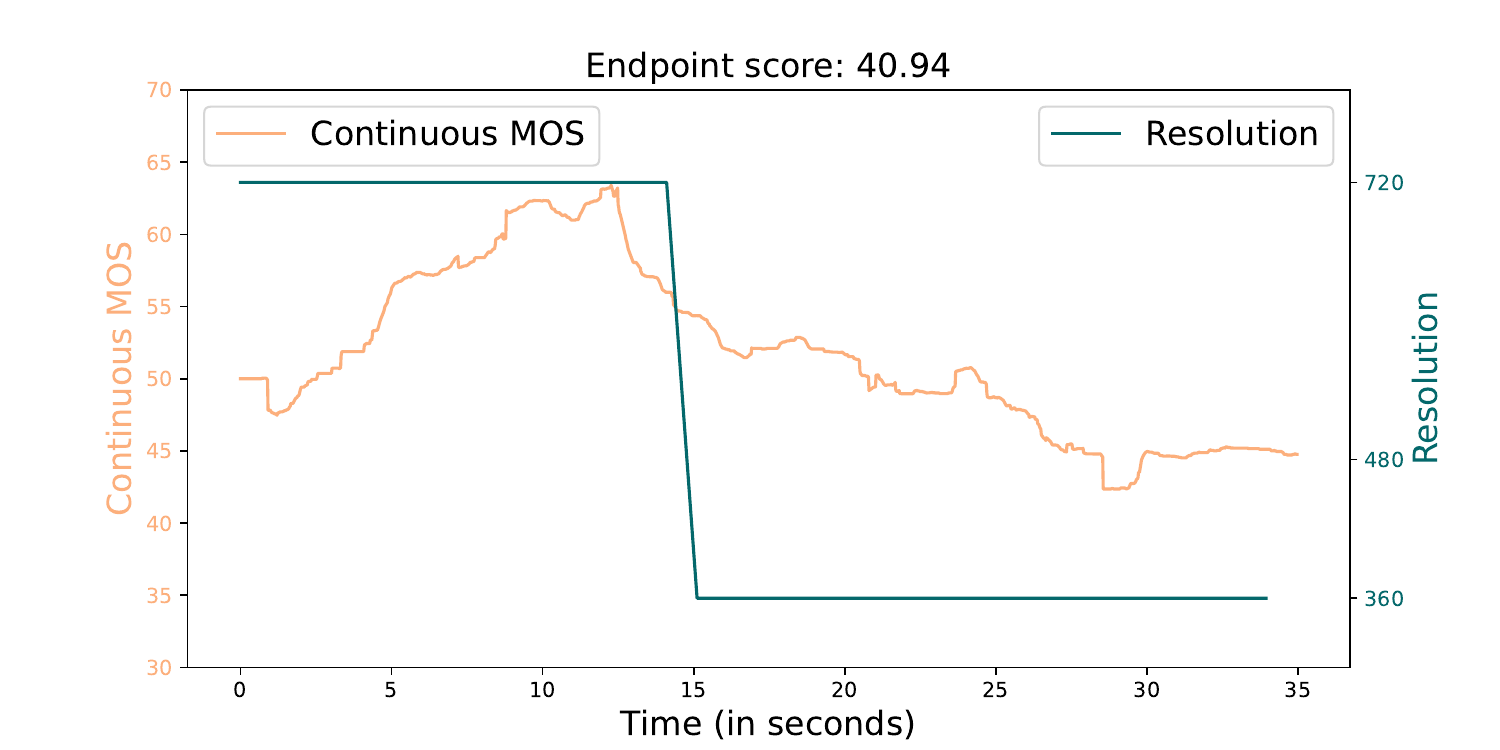}}
\caption{Continuous QoE scores, endpoint QoE scores, and resolutions of 2 pairs of video sequences. (a) and (b) plot the continuous QoE scores (MOS) and time-varying resolutions of two videos drawn from the same source streaming video. (a) shows the resolution decrease from 720p to 480p (green) with corresponding human response in the form of continuous MOS (yellow). (b) shows the continuous QoE score (MOS) in yellow in response to a resolution decrease from 720p to 360p (green). (c) and (d) also plot the continuous MOS (yellow) and changing resolutions (green) of two videos from the same source. (c). A two-step resolution drops from 720p to 480p to 360p. (d) A large one-step resolution drop from 720p to 360p.}
\label{resolution}
\end{figure*}

Resolution significantly influences the perceptual quality of viewed video frames. The videos used in the study ranged from 1080p to 144p in resolution. Among groupings of videos having similar spatial and temporal complexities, increased resolution uniformly resulted in higher MOS. Sudden drops in resolution resulted in corresponding drops in continuous QoE scores, and sharper declines in resolution resulted in more pronounced dips in real-time QoE than more gradual changes, as illustrated in Figs. \ref{resolution} (a)-\ref{resolution} (d). 

However, when the finishing resolutions were similar, whether the resolution decreased in stages suddenly or plunged multiple levels, the real-time QoE scores tended to stabilize around the same value (Fig. \ref{resolution} (c) and \ref{resolution} (d)). Yet, videos that underwent stepwise resolution reductions attained superior endpoint scores. Similar to the behavioral responses to stalls, changes of resolution towards the end of a video playback more heavily impacted the cumulative QoE scores, a clear recency effect. Further, videos having higher spatial and temporal complexities exhibited more pronounced sensitivity to resolution adjustments, which manifested as more significant variations in MOS with changes in resolution.

\subsection{Effect of Bitrate on QoE}

The relationship between bitrate and QoE is complex and multifaceted. Higher bitrates are generally associated with higher video quality, as they allow for a more detailed representation of visual information, thereby enhancing the viewer's experience. For videos having high spatial and temporal complexity—those with detailed visual elements and/or rapid movement—bitrate becomes a critical factor. These types of content generally demand higher bitrates to maintain quality without noticeable degradation. Conversely, on simpler content, increases in bitrate may yield diminishing returns in perceived quality improvements.

In our analysis of streaming performance, we examined the relationship between various statistical measures of bitrate during the streaming process—minimum, maximum, median, average, and standard deviation—and the endpoint QoE scores. Among these measures, the linear correlation between maximum bitrate and the endpoint QoE scores was the most pronounced at 0.35. This finding underscores that while bitrate is a significant determinant of QoE, it is not the only one. Other factors, such as video resolution, encoding efficiency, network stability, and the presence or absence of playback interruptions (stalls), also play essential roles. The overall QoE is the result of the interplay between these elements, with bitrate being an important, but not exclusive, contributor.

\subsection{Effect of Byte Size and Throughput on QoE}
\begin{figure}[!t]
\centering
\subfloat[]{\includegraphics[width=0.25\textwidth]{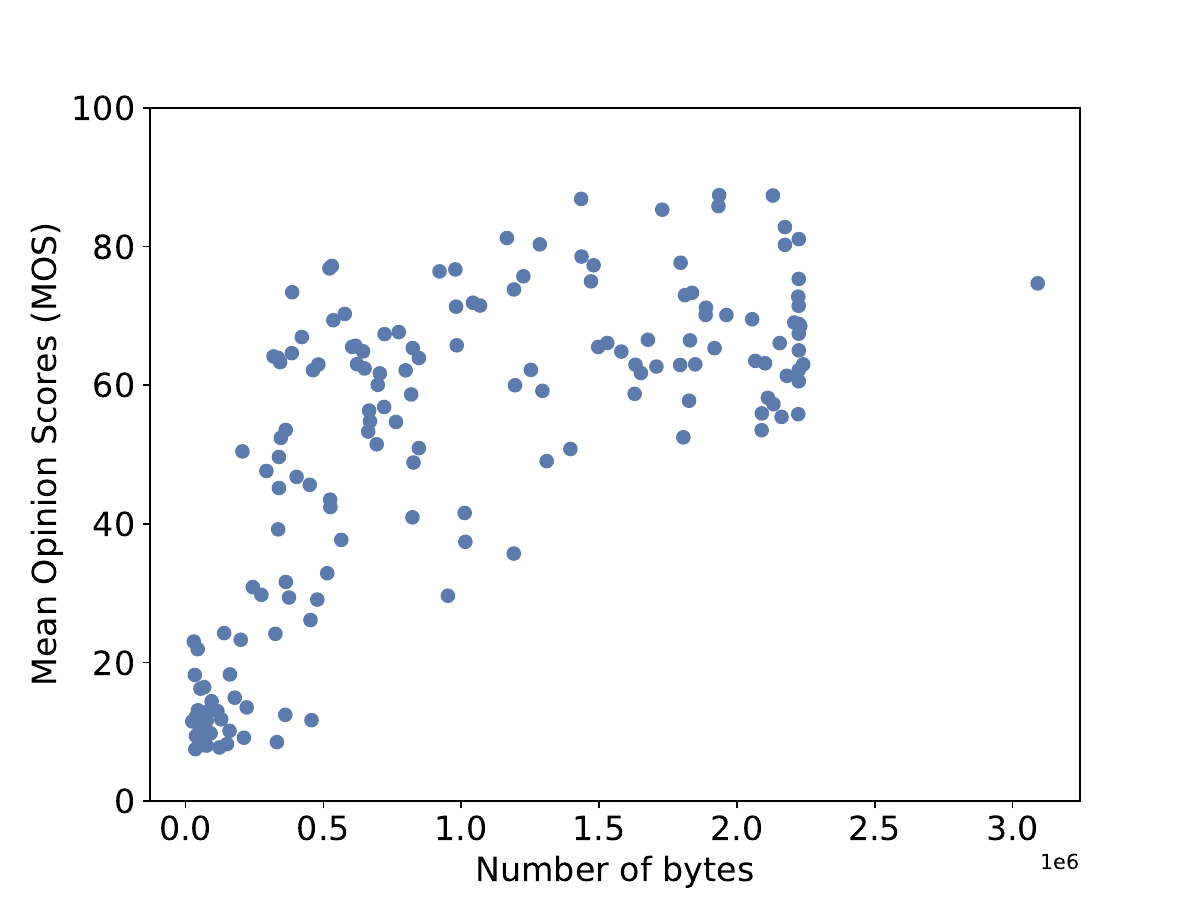}}
\subfloat[]{\includegraphics[width=0.25\textwidth]{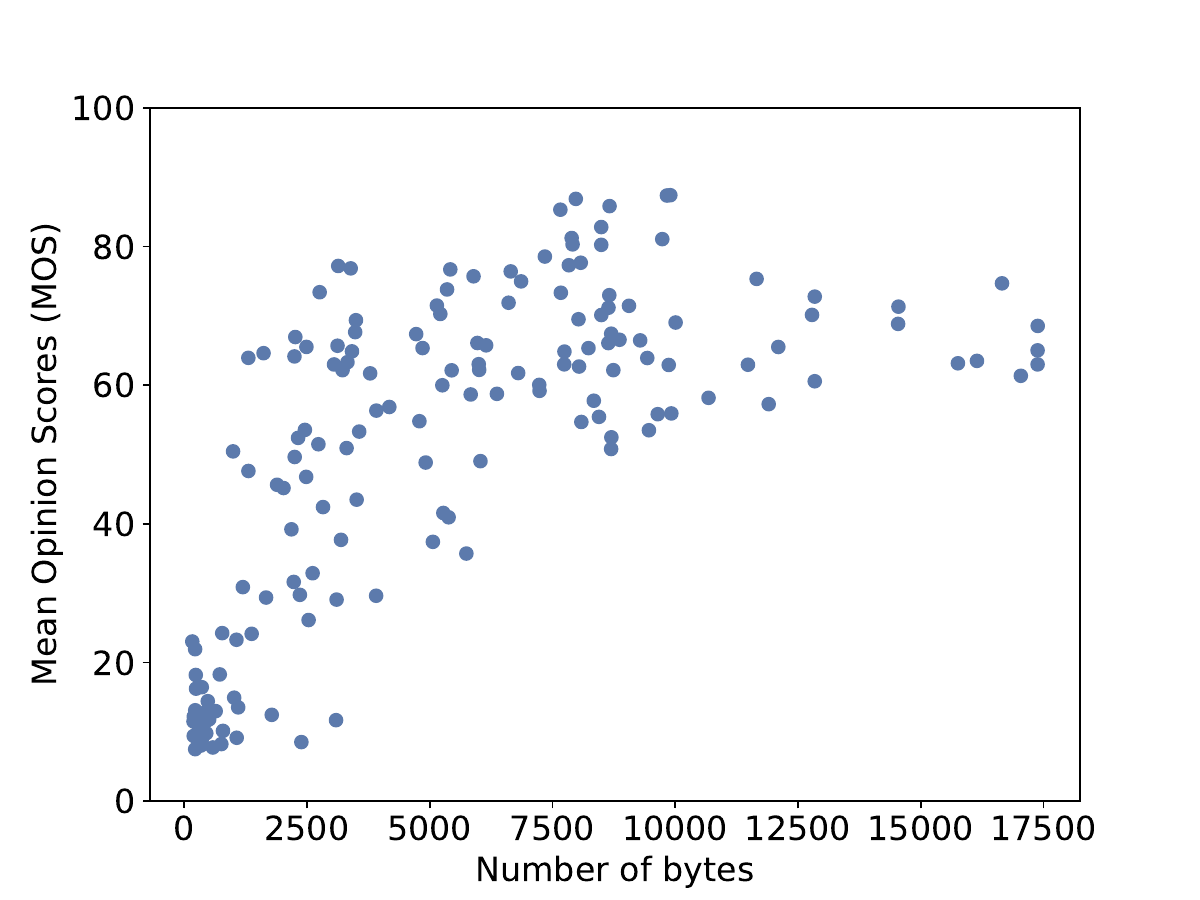}}
\caption{Scatter plots of (a) maximum size of transferred data (bytes) per 2-second interval against endpoint MOS. (b) Scatter plot of peak average throughput (also in bytes) per 2-second interval against endpoint MOS.}
\label{byte_throughput}
\end{figure}

Both transferred data size per second and throughput are critical factors that influence real-time and endpoint QoE during video streaming. By comprehensive analysis of the average, minimum, maximum, standard deviation, and median sizes of transferred data, as well as the average throughput at various intervals throughout the streaming process, it is clear that maintaining an adequate byte size and ensuring stable and sufficient throughput is essential for optimizing both the immediate and overall streaming video experience\cite{ref24}. Delving deeper into the data, we have observed that the maximum size of the transferred data and the peak throughput values have the most significant impact on the endpoint QoE. Fig. \ref{byte_throughput} (a) plots the maximum size of the transferred data (in bytes) against endpoint MOS, while Fig. \ref{byte_throughput} (b) plots the peak throughput (also in bytes) against endpoint MOS. This suggests that spikes in data transfer and throughput are particularly influential in shaping the final evaluation of users' streaming experiences. These spikes represent moments of high-quality streaming, which users come to expect.

Moreover, with regard to continuous QoE scores, there is a noticeable trend whereby user satisfaction—as reported in the form of QoE scores—tends to rise or fall with increases or decreases in the transferred data size and throughput. This relationship, however, is subject to latency periods determined by the buffering techniques used by the client and the time required by human users to perceive and respond to changes in streaming quality.

\subsection{Effect of Idle Transmission Time Ratio}
The Idle Time Percentage is a metric that quantifies the ratio of time intervals over which no data is being transmitted from the streaming server to the local machine, in relation to the overall duration of the video playing. A higher percentage of idle time is indicative of a favorable network condition, as it implies efficient and rapid transmission of data packets or chunks. Conversely, a diminished proportion of idle time could indicate the presence of network congestion, packet loss, or similar complications, which may result in video buffering, diminished video quality, and unsatisfactory viewing experiences. 

\begin{figure}[!t]
\centering
\includegraphics[width=0.3\textwidth]{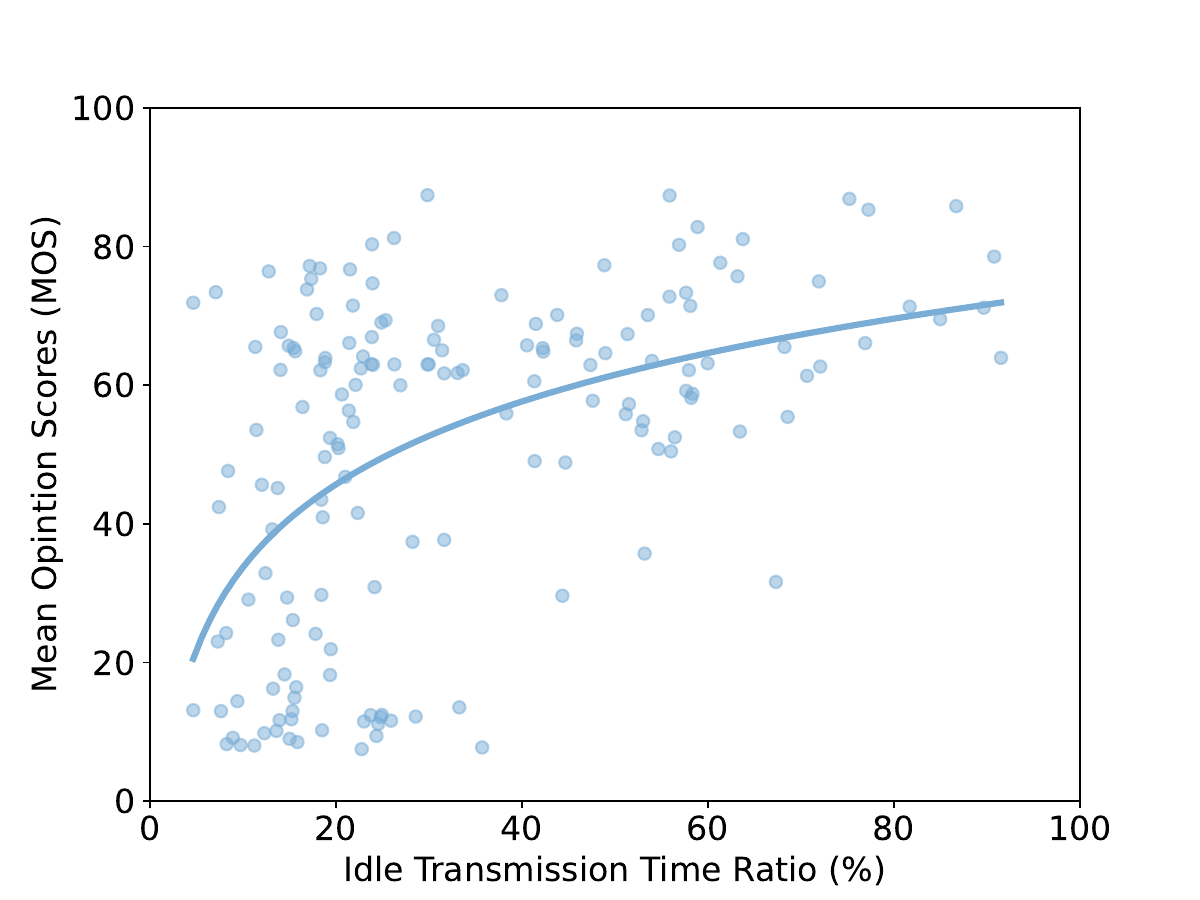}
\caption{Scatter plot of transmission idle time ratio against average endpoint QoE scores.}
\label{idle}
\end{figure}

Fig. \ref{idle} plots the relationship between the measured transmission idle time percentage for each video against its endpoint MOS. As the Idle Transmission Time Ratio increases, an initial significant increase in MOS may be observed, which suggests that up to a certain point, having more idle time (corresponding to faster data transmission and buffer filling) generally leads to better user experiences. Users likely perceive fewer interruptions, less buffering, and smoother playback, which generally leads to higher satisfaction ratings. However, beyond a certain threshold, the increase in MOS with respect to the Idle Transmission Time Ratio starts to plateau. This indicates that after a certain point, the benefit of additional idle time on perceived quality diminishes. It could be that once the buffer is sufficiently ahead of the playback, additional idle time doesn't translate to noticeable improvements in user experience. Users may not perceive a difference between a video that buffers well ahead of time versus one that is excessively buffered.

\section{PERFORMANCE OF OBJECTIVE QoE MODELS}
\label{sec7}
To demonstrate the usefulness of the new database, and also determine how well existing models can be applied for this use case, we studied the performance of both VQA models (which rely on pixel-based video features only) as well as more comprehensive QoE models (which incorporate more features pertaining to video streaming, like stalling).
We also propose a way to utilize network features that are accessible to ISPs to predict QoE without accessing the pixel-domain of compressed domain video features.

\subsection{Existing QoE Models}
\label{ssec:existing-qoe-models}

Because our database is composed of authentically streamed YouTube UGC videos, without access to pristine reference content, we focus on no-reference models in our analysis, including learning-based models, where some of them were retrained by dividing the dataset into two non-overlapping subsets, training set and test set, using an 80\%-20\% split. For the learning-based models BRISQUE \cite{ref35} and Video ATLAS \cite{ref25}, we conducted our experiments by training on 80\% of the dataset and testing their performance on the remaining 20\%. For the other three models, including NIQE \cite{ref34}, SQI \cite{ref10}, and ITU-T Rec. P.1203 model \cite{ref37,ref38}, we evaluated their performance across the entire dataset.

\subsubsection{No-Reference Models}


We evaluated several algorithms, including two no-reference VQA models: NIQE \cite{ref34}, which requires no training, and BRISQUE \cite{ref35}, a trainable model using more features than NIQE. NIQE and BRISQUE are widely used no-reference VQA models that are not suitable for evaluating frame freezes, but were included to establish a baseline. There are also some other models integrating additional streaming factors like buffering and stability: We examined the Streaming QoE Index (SQI) \cite{ref10} and Video ATLAS \cite{ref25}, which combines video presentation quality with the impact of initial buffering and stalling events, offering a more comprehensive assessment of QoE.

\subsubsection{ITU-T Rec. P.1203 Model}

We evaluated the ITU-T Rec. P.1203 QoE model \cite{ref37,ref38}, which was architected to assess the influence of adaptive bitrate (ABR) streaming on streaming video quality. The P.1203 standard describes a modular QoE model consisting of three distinct components. Each of these components is available in a dedicated standard and addresses different parts of the streaming session: per-second video and audio quality (ITU-T Rec. P.1203.1 and P.1203.2), and overall temporal integration considering stalling and quality fluctuations (ITU-T Rec. P.1203.3). Since the original standardized P.1203.1 model does not support the AV1 and VP9 codecs used by YouTube and is limited to a resolution of 1080p at 25 fps, we opted to replace the video part with a more recently published model. The AVQBits$\vert$M0 model~\cite{rao2022p1204extensions,rao2020p1204} was developed as a variant of the P.1204.3 bitstream-based model, but it relies only on metadata like bitrate, codec, resolution, and framerate. In our analysis, the Surfmeter tool that was used to capture the recordings implemented a version of the publicly available AVQBits$\vert$M0 reference code\footnote{\url{https://github.com/Telecommunication-Telemedia-Assessment/p1204_3_extensions}}, with the addition of new model coefficients for the AV1 codec. Since the P.1203 model produces multiple outputs, we used the O.46 (overall MOS) as the value to compare against the subjective scores. Note that for test sequences not having an initial loading delay, an artificial short delay of $0.0001$\,s was inserted to address a known limitation of the model when no initial loading is present. Also, the tested sequences were sometimes only 30\,s long, while the P.1203 standard recommends usage on videos between 1--5\,min only. We have not found this to have any negative impact on the results.

\subsubsection{Performance Analysis}

The model performance metrics utilized were the median Pearson Linear Correlation Coefficient (PLCC), the median Spearman Rank Order Correlation Coefficient (SROCC), and the Root Mean Squared Error (RMSE) over the 100 splits.

\begin{table}[!t]
\caption{Performance of the Overall QoE Predictors on the new LIVE-Viasat Real-World Satellite QoE Database. NIQE and BRISQUE are standard NR-VQA models as baseline comparisons.\label{tab:table4}}
\centering
\begin{tabular}{lrrr}
\toprule
 \thead{Model}     & \thead{SROCC}  & \thead{PLCC} & \thead{RMSE}    \\ 
\midrule
NIQE        & 0.7864 & 0.8019 & 49.5816\\ 
BRISQUE      & 0.6732 & 0.7044 & 21.6653\\ 
SQI         & 0.8029 & 0.8135 & 18.9204\\ 
Video ATLAS & \textbf{0.8962} & 0.9032 & \textbf{11.6085}\\ 
P.1203 with AVQBits$\vert$M0 & 0.8731 & \textbf{0.9424} & 19.1503\\ 
\bottomrule
\end{tabular}
\end{table}

The VQA models, BRISQUE \cite{ref35} and NIQE \cite{ref34}, achieved reasonably high performance, as shown in Table \ref{tab:table4}. This accuracy of traditional VQA models may be attributed to the accurate measurement of distortions \textit{between} instances of stall occurrence. Among the real-world streaming video clips contained in the LIVE-Viasat Real-World Satellite QoE Database, those affected by rebuffering events commonly also suffer from low resolution and/or bitrate. BRISQUE and NIQE effectively account for the perceptual impact of spatial quality degradations while altogether failing to capture stalls, but since stalls frequently coincide with resolution/compression distortions, the predictions they produce are often reasonably accurate.

As shown in Table \ref{tab:table4}, the QoE models proved superior to traditional VQA models as they offer more comprehensive and realistic evaluations of user experience by integrating factors beyond video quality, such as buffering and stream stability. By adding penalties for initial buffering and stalling to the presentation quality, SQI provides a more comprehensive assessment of streaming video QoE. While SQI offers a robust and comprehensive view of the overall streaming experience, Video ATLAS \cite{ref25} employs an integrated approach, offering a comprehensive assessment of QoE. This model is designed to analyze rebuffering events, contributing to its detailed evaluation of streaming experiences. Its integrative approach aligns closely with the multifaceted nature of user experience when viewing diverse streaming video scenarios. Also of note is the ITU-T Rec P.1203 model, which provides a holistic assessment by considering a broad range of factors influencing the streaming experience. Its comprehensive approach is suitable for the complex environments of satellite-based internet services, as a way of predicting the experiences of users under variable streaming conditions. 

\subsection{Mapping Network Measurements to QoE}

Predicting a user's QoE based solely on baseline network parameters, without access to pixels, frames, or compressed domain features is a considerable challenge. This task becomes particularly arduous when analyzing encrypted satellite streams, such as YouTube videos encrypted using the QUIC protocol. Given these constraints, our approach uses carefully selected accessible and reliable network parameters that are mapped to MOS. In our analysis, we used the network parameters from 161 video clips collected using the Viasat satellite network to construct a satellite network-level QoE prediction model. Our baseline feature set includes statistical calculations such as mean, min, max, median, and standard deviation of transfer byte size across various time intervals, along with overall throughput and idle transfer time ratio, which were discussed in the preceding Section \ref{sec6}. Besides the baseline network parameter, we also experimented with the sequences of byte counts received during steaming video playback to evaluate the effect of the temporal features on the users' perception scores. The details of the network parameters are described in Supplementary Material. We used the LIVE-Viasat Real-World Satellite QoE Database to learn mappings from the network measurements to human retrospective QoE scores. We refer to our model as the Baseline Network Measurement (BNM) model, of which we tested three variations.

We conducted 100 trials of each model, employing a cross-validation technique with random 80-20 splits into training and test content. This approach ensured that no content overlapped between the training and test sets, thereby eliminating content bias. We utilized four different regression models to predict MOS: Multi-Linear Regression, Decision Tree, Random Forest, and Support Vector Machine (SVR) to learn to predict the MOS. To enhance our assessment of model performance beyond using the BNM, we analyzed the sequences of byte counts received by the local machine over both one-second and three-second intervals during video streaming playback, aiming to understand how these temporal data features influenced users' perception scores. To further refine our models, we employed GridSearch to tune the hyperparameters of the Decision Tree, Random Forest, and SVR models. This process was conducted on each of the 100 random 80-20 splits, ensuring that the models were optimized for each individual training set. For each model and different feature set, we report the median SROCC, PLCC and RMSE over 100 random trials. A comparative analysis of models that integrated these temporal features and those that did not (the baseline set) is shown in Table \ref{tab:table6}. Table \ref{tab:table6} shows that all three versions of BNM models perform quite well, nearly matching the performance of the models in Table \ref{tab:table4}, all of which make use of much richer information.

Upon analyzing the performance of the tested model, we observed that high-quality QoE predictions can be achieved without direct QoE measurements on the video contents, such as pixel information, framerates, or compression metrics. This is particularly remarkable given that the learned QoE analyzers rely only on network parameters, which are often considered less informative than direct media quality indicators. The SVR model demonstrated the best prediction capabilities, followed by Random Forest, Multi-Linear Regression, and lastly, the Decision Tree. This ordering underscores the SVR's ability to manage complex, nonlinear relationships within high-dimensional data. However, even the Multi-Linear Regression model delivered a respectable performance. Overall, these results highlight the robustness of our feature selection process.

The incremental improvements given by the inclusion of byte count sequences per one-second and three-second intervals in the Random Forest and SVR models, in particular, reveal that temporal features possess important information regarding user experience. This nuanced enhancement of all SROCC, PLCC and RMSE for models leveraging these additional temporal features suggests that the complexity of models like Random Forest and SVR can indeed capture and utilize subtle temporal patterns that correlate with users' perceived QoE.  While the Multi-Linear Regression and Decision Tree models did not yield significant gains with the inclusion of temporal data, the Random Forest and SVR's ability to exploit temporal features is indicative both of the importance of temporal features and of choosing an effective regressor when predicting user QoE. 

\begin{table*}[!tb]

  \caption{Performances of Endpoint QoE predictors using different sets of network measurements, mapped to MOS by four different regression models. The best performance of the evaluation model is boldfaced. SVR = Support Vector Regression, MLR = Multi-Linear Regression, DT = Decision Tree, RF = Random Forest.}
  \label{tab:table6}
  \centering
\resizebox{\textwidth}{!}{
  \begin{tabular}{lcccccccccccc}
  \toprule
   & \multicolumn{4}{c}{\textbf{SROCC}} & \multicolumn{4}{c}{\textbf{PLCC}} & \multicolumn{4}{c}{\textbf{RSME}}\\
  \cmidrule(lr){2-5} \cmidrule(lr){6-9}  \cmidrule(lr){10-13}
  \textbf{Feature Set} & \textbf{MLR} & \textbf{DT} & \textbf{RF} & \textbf{SVR} & \textbf{MLR} & \textbf{DT} & \textbf{RF} & \textbf{SVR} & \textbf{MLR} & \textbf{DT} & \textbf{RF} & \textbf{SVR} \\
  \midrule
  Baseline Network Measurements & 0.7289 & 0.7175 & 0.7530 & 0.7782 & 0.7245 & 0.7023 & 0.7332 & 0.8482 & 19.5358 & 14.9976 & 12.1259 & 11.7775\\
  Baseline Network Measurements\\+ bytes/sec & 0.7202 & 0.7088 & 0.7648 & 0.7928 & 0.7014 & 0.6876 & 0.7248 & 0.8457 & 20.9134 & 13.7985 & 11.8142 & 11.7962\\
  Baseline Network Measurements\\+ bytes/sec + bytes/3 sec & 0.7137 & 0.7158 & 0.7680 & \textbf{0.7935} & 0.6834 & 0.6714 & 0.7416 & \textbf{0.8623} & 20.5396 & 13.5968 & 11.7635 & \textbf{11.6247}\\
  \bottomrule
  \end{tabular}}
\end{table*}

\section{Conclusion}

There is a growing need for tools and resources that can enable a better understanding of satellite streaming video QoE under real-world conditions. In particular, there is a dearth of public databases that model real-world QoE-related satellite video distortions. Our database provides a diverse set of 179 streaming satellite videos, related QoE measurements and network packets, all of which were captured from an operational real-world satellite network. We aim to provide insights into the various challenges encountered during video streaming over fluctuating network conditions and to offer valuable data for model-building through these videos and our subjective video QoE study. The associated subjective scores, which we are making publicly available along with the video data and network data will allow other researchers to delve deeper into the nuances of QoE and to develop more powerful and practical satellite video QoE models. To demonstrate the value of the new LIVE-Viasat Real-World Satellite QoE Database, we used it to evaluate both existing VQA and video QoE models. We found that VQA models performed reasonably well, but that video QoE models performed much better as they could account for the effects of both space-time video quality as well as stalls. We also developed simple pixel-free video QoE models that operate only on satellite network-level data and showed that good performance can also be obtained.

\bibliography{viasat_paper_draft}

\begin{thebibliography}{10}
\providecommand{\url}[1]{#1}
\csname url@samestyle\endcsname
\providecommand{\newblock}{\relax}
\providecommand{\bibinfo}[2]{#2}
\providecommand{\BIBentrySTDinterwordspacing}{\spaceskip=0pt\relax}
\providecommand{\BIBentryALTinterwordstretchfactor}{4}
\providecommand{\BIBentryALTinterwordspacing}{\spaceskip=\fontdimen2\font plus
\BIBentryALTinterwordstretchfactor\fontdimen3\font minus \fontdimen4\font\relax}
\providecommand{\BIBforeignlanguage}[2]{{%
\expandafter\ifx\csname l@#1\endcsname\relax
\typeout{** WARNING: IEEEtran.bst: No hyphenation pattern has been}%
\typeout{** loaded for the language `#1'. Using the pattern for}%
\typeout{** the default language instead.}%
\else
\language=\csname l@#1\endcsname
\fi
#2}}
\providecommand{\BIBdecl}{\relax}
\BIBdecl

\bibitem{barnett2018cisco}
T.~Barnett, S.~Jain, U.~Andra, and T.~Khurana, ``Cisco visual networking index (vni) complete forecast update, 2017--2022,'' \emph{Americas/EMEAR Cisco Knowledge Network (CKN) Presentation}, pp. 1--30, 2018.

\bibitem{ref1}
\BIBentryALTinterwordspacing
T.~Stockhammer, ``{Dynamic Adaptive Streaming} over {HTTP}: {Standards and Design Principles},'' in \emph{Proceedings of the Second Annual ACM Conference on Multimedia Systems}.\hskip 1em plus 0.5em minus 0.4em\relax Association for Computing Machinery, 2011, p. 133–144. [Online]. Available: \url{https://doi.org/10.1145/1943552.1943572}
\BIBentrySTDinterwordspacing

\bibitem{ref2}
R.~Pantos and W.~May, ``{HTTP} live streaming,'' 2017.

\bibitem{ref3}
Z.~Wang, A.~Bovik, H.~Sheikh, and E.~Simoncelli, ``Image quality assessment: from error visibility to structural similarity,'' \emph{IEEE Transactions on Image Processing}, vol.~13, no.~4, pp. 600--612, 2004.

\bibitem{ref4}
Z.~Wang, E.~Simoncelli, and A.~Bovik, ``Multiscale structural similarity for image quality assessment,'' in \emph{The Asilomar Conference on Signals, Systems \& Computers}, 2003, pp. 1398--1402 Vol.2.

\bibitem{ref5}
R.~Soundararajan and A.~C. Bovik, ``Video quality assessment by reduced reference spatio-temporal entropic differencing,'' \emph{IEEE Transactions on Circuits and Systems for Video Technology}, vol.~23, no.~4, pp. 684--694, 2012.

\bibitem{ref6}
N.~TechBlog. (2017) {Netflix Technology Blog. Toward} a practical perceptual video quality metric.

\bibitem{ref7}
A.~K. Moorthy, L.~K. Choi, A.~C. Bovik, and G.~de~Veciana, ``{Video Quality Assessment on Mobile Devices: Subjective, Behavioral and Objective Studies},'' \emph{IEEE Journal of Selected Topics in Signal Processing}, vol.~6, no.~6, pp. 652--671, 2012.

\bibitem{ref8}
C.~Chen, L.~K. Choi, G.~de~Veciana, C.~Caramanis, R.~W. Heath, and A.~C. Bovik, ``{Modeling the Time—Varying Subjective Quality of HTTP Video Streams With Rate Adaptations},'' \emph{IEEE Transactions on Image Processing}, vol.~23, no.~5, pp. 2206--2221, 2014.

\bibitem{ref9}
D.~Ghadiyaram, A.~C. Bovik, H.~Yeganeh, R.~Kordasiewicz, and M.~Gallant, ``Study of the effects of stalling events on the quality of experience of mobile streaming videos,'' in \emph{2014 IEEE Global Conference on Signal and Information Processing (GlobalSIP)}, 2014, pp. 989--993.

\bibitem{ref10}
Z.~Duanmu, K.~Zeng, K.~Ma, A.~Rehman, and Z.~Wang, ``{A Quality-of-Experience Index for Streaming Video},'' \emph{IEEE Journal of Selected Topics in Signal Processing}, vol.~11, no.~1, pp. 154--166, 2017.

\bibitem{ref11}
Z.~Duanmu, K.~Ma, and Z.~Wang, ``{Quality-of-Experience of Adaptive Video Streaming: Exploring the Space of Adaptations},'' in \emph{Proceedings of the 25th ACM International Conference on Multimedia}, 2017, pp. 1752--1760.

\bibitem{ref12}
C.~G. Bampis, Z.~Li, A.~K. Moorthy, I.~Katsavounidis, A.~Aaron, and A.~C. Bovik, ``{Study of Temporal Effects on Subjective Video Quality of Experience},'' \emph{IEEE Transactions on Image Processing}, vol.~26, no.~11, pp. 5217--5231, 2017.

\bibitem{ref13}
D.~Ghadiyaram, J.~Pan, and A.~C. Bovik, ``{A Subjective and Objective Study of Stalling Events in Mobile Streaming Videos},'' \emph{IEEE Transactions on Circuits and Systems for Video Technology}, vol.~29, no.~1, pp. 183--197, 2019.

\bibitem{ref14}
\BIBentryALTinterwordspacing
C.~G. Bampis and A.~C. Bovik, ``{Learning to Predict Streaming Video QoE: Distortions, Rebuffering and Memory},'' \emph{CoRR}, vol. abs/1703.00633, 2017. [Online]. Available: \url{http://arxiv.org/abs/1703.00633}
\BIBentrySTDinterwordspacing

\bibitem{ref15}
Z.~Duanmu, A.~Rehman, and Z.~Wang, ``{A Quality-of-Experience Database for Adaptive Video Streaming},'' \emph{IEEE Transactions on Broadcasting}, vol.~64, no.~2, pp. 474--487, 2018.

\bibitem{ref16}
\BIBentryALTinterwordspacing
Z.~Duanmu, W.~Liu, Z.~Li, D.~Chen, Z.~Wang, Y.~Wang, and W.~Gao, ``{Assessing the Quality-of-Experience of Adaptive Bitrate Video Streaming},'' \emph{ArXiv}, vol. abs/2008.08804, 2020. [Online]. Available: \url{https://api.semanticscholar.org/CorpusID:221186589}
\BIBentrySTDinterwordspacing

\bibitem{ref17}
C.~G. Bampis, Z.~Li, I.~Katsavounidis, T.-Y. Huang, C.~Ekanadham, and A.~C. Bovik, ``{Towards Perceptually Optimized Adaptive Video Streaming-A Realistic Quality of Experience Database},'' \emph{IEEE Transactions on Image Processing}, vol.~30, pp. 5182--5197, 2021.

\bibitem{ref18}
D.~Ghadiyaram, J.~Pan, and A.~C. Bovik, ``{Learning a Continuous-Time Streaming Video QoE Model},'' \emph{IEEE Transactions on Image Processing}, vol.~27, no.~5, pp. 2257--2271, 2018.

\bibitem{ref19}
M.~Gadaleta, F.~Chiariotti, M.~Rossi, and A.~Zanella, ``{D-DASH: A Deep Q-Learning Framework for DASH Video Streaming},'' \emph{IEEE Transactions on Cognitive Communications and Networking}, vol.~3, no.~4, pp. 703--718, 2017.

\bibitem{ref20}
T.~Mangla, E.~Halepovic, M.~Ammar, and E.~Zegura, ``{eMIMIC: Estimating HTTP-based video QoE metrics from encrypted network traffic},'' in \emph{IEEE Network Traffic Measurement and Analysis Conference (TMA)}, 2018, pp. 1--8.

\bibitem{ref21}
V.~Krishnamoorthi, N.~Carlsson, E.~Halepovic, and E.~Petajan, ``{BUFFEST: Predicting buffer conditions and real-time requirements of HTTP(s) adaptive streaming clients},'' pp. 76--87, 2017.

\bibitem{ref22}
V.~Aggarwal, E.~Halepovic, J.~Pang, S.~Venkataraman, and H.~Yan, ``{Prometheus: Toward quality-of-experience estimation for mobile apps from passive network measurements},'' in \emph{Proceedings of the 15th Workshop on Mobile Computing Systems and Applications}, 2014, pp. 1--6.

\bibitem{ref23}
G.~Dimopoulos, I.~Leontiadis, P.~Barlet-Ros, and K.~Papagiannaki, ``Measuring video {QoE} from encrypted traffic,'' in \emph{Proceedings of the 2016 Internet Measurement Conference}, 2016, pp. 513--526.

\bibitem{ref24}
I.~Orsolic, D.~Pevec, M.~Suznjevic, and L.~Skorin-Kapov, ``{A machine learning approach to classifying YouTube QoE based on encrypted network traffic},'' \emph{Multimedia tools and applications}, vol.~76, pp. 22\,267--22\,301, 2017.

\bibitem{ref25}
D.~Tsilimantos, T.~Karagkioules, and S.~Valentin, ``{Classifying flows and buffer state for YouTube's HTTP adaptive streaming service in mobile networks},'' in \emph{Proceedings of the 9th ACM Multimedia Systems Conference}, 2018, pp. 138--149.

\bibitem{ref26}
M.~H. Mazhar and Z.~Shafiq, ``{Real-time video quality of experience monitoring for HTTPs and QUIC},'' in \emph{IEEE INFOCOM Conference on Computer Communications}, 2018, pp. 1331--1339.

\bibitem{ref28}
Z.~Li and C.~G. Bampis, ``Recover subjective quality scores from noisy measurements,'' in \emph{IEEE Data Compression Conference (DCC)}.\hskip 1em plus 0.5em minus 0.4em\relax IEEE, 2017, pp. 52--61.

\bibitem{ref29}
Z.~Li, C.~G. Bampis, L.~Krasula, L.~Janowski, and I.~Katsavounidis, ``A simple model for subject behavior in subjective experiments,'' \emph{arXiv preprint arXiv:2004.02067}, 2020.

\bibitem{ref30}
A.~M. Van~Dijk, J.-B. Martens, and A.~B. Watson, ``Quality assessment of coded images using numerical category scaling,'' in \emph{Advanced Image and Video Communications and Storage Technologies}, vol. 2451, 1995, pp. 90--101.

\bibitem{ref31}
D.~J. Berndt and J.~Clifford, ``Using dynamic time warping to find patterns in time series,'' in \emph{The 3rd International Conference on Knowledge Discovery and Data Mining}, 1994, pp. 359--370.

\bibitem{ref32}
M.~Hubert and E.~Vandervieren, ``{An Adjusted Boxplot for Skewed Distributions},'' \emph{Computational Statistics and Data Analysis}, vol.~52, pp. 5186--5201, 08 2008.

\bibitem{ref33}
K.~Seshadrinathan and A.~C. Bovik, ``Temporal hysteresis model of time varying subjective video quality,'' in \emph{IEEE International Conference on Acoustics, Speech and Signal Processing (ICASSP)}, 2011, pp. 1153--1156.

\bibitem{ref34}
A.~Mittal, R.~Soundararajan, and A.~C. Bovik, ``Making a “completely blind” image quality analyzer,'' \emph{IEEE Signal Processing Letters}, vol.~20, no.~3, pp. 209--212, 2012.

\bibitem{ref35}
A.~Mittal, A.~K. Moorthy, and A.~C. Bovik, ``No-reference image quality assessment in the spatial domain,'' \emph{IEEE Transactions on Image Processing}, vol.~21, no.~12, pp. 4695--4708, 2012.

\bibitem{ref37}
\BIBentryALTinterwordspacing
A.~Raake, M.-N. Garcia, W.~Robitza, P.~List, S.~Göring, and B.~Feiten, ``{A bitstream-based, scalable video-quality model for HTTP adaptive streaming: ITU-T P.1203.1},'' in \emph{Ninth International Conference on Quality of Multimedia Experience (QoMEX)}.\hskip 1em plus 0.5em minus 0.4em\relax Erfurt: IEEE, May 2017. [Online]. Available: \url{http://ieeexplore.ieee.org/document/7965631/}
\BIBentrySTDinterwordspacing

\bibitem{ref38}
W.~Robitza, S.~Göring, A.~Raake, D.~Lindegren, G.~Heikkilä, J.~Gustafsson, P.~List, B.~Feiten, U.~Wüstenhagen, M.-N. Garcia, K.~Yamagishi, and S.~Broom, ``{HTTP Adaptive Streaming QoE Estimation with ITU-T Rec. P.1203 – Open Databases and Software},'' in \emph{9th ACM Multimedia Systems Conference}, Amsterdam, 2018.

\bibitem{rao2022p1204extensions}
R.~R. Ramachandra~Rao, S.~Göring, and A.~Raake, ``Avqbits—adaptive video quality model based on bitstream information for various video applications,'' \emph{IEEE Access}, vol.~10, pp. 80\,321--80\,351, 2022.

\bibitem{rao2020p1204}
R.~R. {Ramachandra Rao}, S.~G\"oring, W.~Robitza, A.~Raake, B.~Feiten, P.~List, and U.~Wüstenhagen, ``Bitstream-based model standard for 4k/uhd: Itu-t p.1204.3 -- model details, evaluation, analysis and open source implementation,'' in \emph{2020 Twelfth International Conference on Quality of Multimedia Experience (QoMEX)}, Athlone, Ireland, May 2020.

\bibitem{garcia2014quality}
M.-N. Garcia, F.~De~Simone, S.~Tavakoli, N.~Staelens, S.~Egger, K.~Brunnstr{\"o}m, and A.~Raake, ``Quality of experience and http adaptive streaming: A review of subjective studies,'' in \emph{2014 sixth international workshop on quality of multimedia experience (qomex)}.\hskip 1em plus 0.5em minus 0.4em\relax IEEE, 2014, pp. 141--146.

\bibitem{weiss2014temporal}
B.~Weiss, D.~Guse, S.~M{\"o}ller, A.~Raake, A.~Borowiak, and U.~Reiter, ``Temporal development of quality of experience,'' \emph{Quality of Experience: advanced concepts, applications and methods}, pp. 133--147, 2014.

\bibitem{itu-t}
\BIBentryALTinterwordspacing
``{ITU-T P.10/G.100},'' 2017. [Online]. Available: \url{https://www.itu.int/rec/T-REC-P.10-201906-I!Amd1/en}
\BIBentrySTDinterwordspacing

\bibitem{Chakravarty1995MethodologyFT}
\BIBentryALTinterwordspacing
S.~Chakravarty, ``Methodology for the subjective assessment of the quality of television pictures,'' 1995. [Online]. Available: \url{https://api.semanticscholar.org/CorpusID:6155668}
\BIBentrySTDinterwordspacing

\bibitem{itu-t2}
``{ITU-R Rec.BT.500-11},'' 2002.

\bibitem{7327186}
D.~Ghadiyaram and A.~C. Bovik, ``Massive online crowdsourced study of subjective and objective picture quality,'' \emph{IEEE Transactions on Image Processing}, vol.~25, no.~1, pp. 372--387, 2016.

\bibitem{7932975}
D.~Ghadiyaram, J.~Pan, A.~C. Bovik, A.~K. Moorthy, P.~Panda, and K.-C. Yang, ``In-capture mobile video distortions: A study of subjective behavior and objective algorithms,'' \emph{IEEE Transactions on Circuits and Systems for Video Technology}, vol.~28, no.~9, pp. 2061--2077, 2018.

\bibitem{icip}
B.~Chen, Z.~Shang, J.~W. Chung, D.~Lerner, and A.~C. Bovik, ``A real-world satellite video subjective {Q}o{E} database,'' in \emph{2024 IEEE International Conference on Image Processing (ICIP)}.\hskip 1em plus 0.5em minus 0.4em\relax IEEE, 2024, pp. 786--790.

\end{thebibliography}
\bibliographystyle{IEEEtran}



\end{document}